\documentclass[journal]{IEEEtran}
\usepackage[T1]{fontenc}
\usepackage{cite}
\usepackage{amsmath,amssymb}
\usepackage{graphicx}
\usepackage[table]{xcolor}
\usepackage{booktabs,tabularx,array,longtable}
\usepackage{multirow}
\usepackage{threeparttable}
\usepackage{stfloats}
\usepackage{multicol}
\usepackage{placeins}
\usepackage{balance}
\usepackage{hyperref}
\usepackage{orcidlink}

\title{Earth Observation Foundation Models for Terrestrial Ecohydrology: From Representation Learning to Process Inference}

\author{Yi~Yu\orcidlink{0000-0002-1140-2713}, 
        Jian~Peng\orcidlink{0000-0002-4071-0512},
        Yucheng~Lin\orcidlink{0000-0002-5556-4490},
        Trevor F. Keenan\orcidlink{0000-0002-3347-0258}
        and Thomas~F.~A.~Bishop\orcidlink{0000-0002-6723-7323}

        \thanks{This work was supported in part by the University of Sydney through the SOLES Strategic Partnership Seeding Grant, and in part by a Google Cloud Research Credits Grant. \textit{Corresponding author: Yi Yu.}}
        \thanks{Yi Yu and Thomas F. A. Bishop are with the Precision Agriculture, Hydrology and Geoinformation Science Laboratory, School of Life and Environmental Sciences, The University of Sydney, Eveleigh, NSW, Australia (e-mail: yi.yu1@sydney.edu.au; thomas.bishop@sydney.edu.au).}
        \thanks{Jian Peng is with the Department of Remote Sensing, Helmholtz Centre for Environmental Research--UFZ, Leipzig, Germany, and also with the Institute for Earth System Science and Remote Sensing, Leipzig University, Leipzig, Germany (e-mail: jian.peng@ufz.de).}
        \thanks{Yucheng Lin is with the School of Energy and Environment, City University of Hong Kong, Hong Kong SAR, China (e-mail: yucheng.lin@cityu.edu.hk).}
        \thanks{Trevor F. Keenan is with the Department of Environmental Science, Policy, and Management, University of California, Berkeley, Berkeley, CA, USA, and also with the Climate and Ecosystem Sciences Division, Lawrence Berkeley National Laboratory, Berkeley, CA, USA (e-mail: trevorkeenan@berkeley.edu).}
}

\begin{document}

\maketitle

\begin{abstract}

Earth observation foundation models (EOFMs) are emerging as reusable representation frameworks for data-driven retrieval, prediction and process modelling within ecohydrology, which integrate EO, meteorological forcing and process models to characterise coupled water, energy and carbon dynamics in vegetation and soil across scales. However, there is yet to be an ecohydrology-specific synthesis assessing the EOFM relevance, application evidence or evaluation requirements under uncertain reference data, scale mismatch, temporal dependence and distribution shift. Here, we develop a framework for determining when EOFMs support scientifically interpretable inference, and identify a mismatch between current EOFM development and the requirements of ecohydrological process inference. Firstly, we establish an observation-to-inference hierarchy, demonstrating that EOFM relevance depends on target-specific sensing pathways, spatial and temporal support, and traceable uncertainty.
Secondly, we conduct a meta-analysis, which reveals that pretraining is dominated by reflected optical and active-microwave data, with sparse thermal coverage and no explicit passive-microwave-emission or SIF sensing data sources. The heterogeneous targets and validation designs of EOFMs motivate an application-level assessment of demonstrated capabilities and evidentiary strength. Thirdly, we synthesise the EOFM application records for ecohydrology, and find the strongest support for spatial context, label-efficient adaptation and hybrid workflows; evidence weakens with inference depth, and independent validation of fluxes, coupled dynamics, event trajectories, calibrated uncertainty and decision benefit remains sparse. Fourthly, our audit of benchmark studies shows stronger coverage of fair adaptation and reproducibility in general EOFM suites and of process targets, direct reference evidence and distribution shifts in ecohydrological evaluations; physical consistency and uncertainty remain weakly covered. Taken together, these findings motivate a process-aware framework that aligns EOFM design and evaluation with the inferred ecohydrological variable, observation pathway and process timescale, supporting trustworthy monitoring and interpretation of coupled water, energy and carbon dynamics.

\end{abstract}


\section{Introduction}

\IEEEPARstart{E}{cohydrology} studies how water availability and movement interact with soils, vegetation and the atmosphere to shape ecological processes across space and time \cite{rodriguez-iturbe_ecohydrology_2000,seneviratne_investigating_2010}. These interactions determine how precipitation and stored soil water are partitioned among runoff, evaporation and transpiration; how available energy is divided between sensible and latent heat; and how stomatal regulation couples plant water loss with photosynthetic carbon uptake \cite{seneviratne_investigating_2010,gentine_coupling_2019}. Understanding these processes is essential for explaining ecosystem productivity and responses to climate variability and drought, and for quantifying feedback between terrestrial water, energy and carbon cycles \cite{seneviratne_investigating_2010,gentine_coupling_2019}. For terrestrial ecohydrology, Earth observation (EO) refers to systematic measurements of the land surface, vegetation and atmosphere acquired by satellite, airborne and in situ sensors \cite{li_regulation_2024,jiao_remote_2026}. These complementary EO data provide information on vegetation condition \cite{zeng_optical_2022}, surface energy exchange \cite{fisher_evapotranspiration_2026}, soil moisture (SM) \cite{peng_roadmap_2021}, vegetation water content (VWC) \cite{ma_satellite_2025} and photosynthetic functioning \cite{mohammed_remote_2019} across space and time. Together with meteorological forcing and process models, these complementary EO data connect local process understanding with regional-to-global analyses of SM dry-down \cite{dorigo_international_2011}, drought onset and recovery \cite{mishra_review_2010}, land--atmosphere energy exchange \cite{pastorello_fluxnet2015_2020}, plant water use \cite{li_regulation_2024} and carbon uptake \cite{jung_fluxcom_2019}. Their scale, heterogeneity and temporal detail consequently increase the need to integrate multiple EO sources and resolve complex ecohydrological relationships.

\begin{table*}[!b]
\centering
\begin{threeparttable}
\caption{Representative reviews and surveys of FMs in RS, EO and geoscience. Binary scope is coded at the EO data, model-integration and evaluation levels. A check mark requires substantive coverage through a dedicated section or subsection, a comparison table or an original comparative experiment; a cross denotes incidental or absent coverage.}
\label{tab:eofm_review_scope}
\footnotesize
\setlength{\tabcolsep}{2.5pt}
\renewcommand{\arraystretch}{1.15}
\begin{tabularx}{\textwidth}{>{\raggedright\arraybackslash}p{0.105\textwidth} >{\centering\arraybackslash}p{0.045\textwidth} >{\raggedright\arraybackslash}p{0.06\textwidth} >{\raggedright\arraybackslash}X *{6}{>{\centering\arraybackslash}p{0.069\textwidth}}}
\toprule
Publication & Year & Journal & Primary scope and coverage & \multicolumn{2}{c}{EO data coverage} & \multicolumn{2}{c}{Model integration} & \multicolumn{2}{c}{Evaluation} \\
\cmidrule(lr){5-6}\cmidrule(lr){7-8}\cmidrule(lr){9-10}
 & & & & \shortstack{EO\\modalities} & \shortstack{Time\\series} & \shortstack{Adaptation/\\transfer} & \shortstack{Domain\\knowledge} & \shortstack{Comparative\\benchmarks} & \shortstack{Process\\targets} \\
\midrule
Jiao et al. \cite{jiao_brain-inspired_2023} & 2023 & JSTARS & RS FMs, downstream classification, localisation and understanding, and a brain-inspired framework & $\checkmark$ & $\checkmark$ & $\checkmark$ & $\checkmark$ & $\checkmark$ & $\times$ \\
Lu et al. \cite{lu_vision_2025} & 2025 & GRSM & Vision foundation-model architectures, pretraining data and methods, performance comparisons and research directions & $\checkmark$ & $\checkmark$ & $\times$ & $\times$ & $\checkmark$ & $\checkmark$ \\
Huo et al. \cite{huo_when_2025} & 2025 & RS & General and domain-specific vision, prompted and heterogeneous FMs for RS & $\checkmark$ & $\checkmark$ & $\checkmark$ & $\checkmark$ & $\checkmark$ & $\times$ \\
Xiao et al. \cite{xiao_foundation_2025} & 2025 & GRSM & Visual, vision--language and language FMs, EO datasets, technical advances and benchmarks & $\checkmark$ & $\checkmark$ & $\checkmark$ & $\checkmark$ & $\checkmark$ & $\times$ \\
Zhang et al. \cite{zhang_when_2025} & 2025 & GRSM & FMs for general geoscientific AI, including vision, language, multimodal and agentic systems & $\checkmark$ & $\times$ & $\checkmark$ & $\checkmark$ & $\times$ & $\checkmark$ \\
Zhou et al. \cite{zhou_advances_2025} & 2025 & RS & Multimodal RS FMs, vision--X pretraining data, architectures and fusion strategies & $\checkmark$ & $\checkmark$ & $\checkmark$ & $\times$ & $\checkmark$ & $\times$ \\
Hong et al. \cite{hong_foundation_2026} & 2026 & GRSM & Technical evolution of RS FMs from unimodal to multimodal learning & $\checkmark$ & $\checkmark$ & $\checkmark$ & $\times$ & $\checkmark$ & $\checkmark$ \\
Ours & -- & -- & EOFMs across ecohydrological inference levels, with physical-consistency and transfer evaluation & $\checkmark$ & $\checkmark$ & $\checkmark$ & $\checkmark$ & $\checkmark$ & $\checkmark$ \\
\bottomrule
\end{tabularx}
\begin{tablenotes}[flushleft]
\footnotesize
\item[] \emph{Abbreviations}: GRSM (IEEE Geoscience and Remote Sensing Magazine); JSTARS (IEEE Journal of Selected Topics in Applied Earth Observations and Remote Sensing); RS (Remote Sensing).
\item[] \emph{Process targets}: Substantive coverage of continuous biophysical states or fluxes, dynamic Earth-system processes or process-linked outcomes.
\end{tablenotes}
\end{threeparttable}
\end{table*}

Data-driven approaches, including both machine learning (ML) and deep learning (DL), have expanded the use of EO data in ecohydrology by learning nonlinear and interacting spatiotemporal relationships for retrieval, prediction, gap filling, data fusion and flux upscaling \cite{zhu_deep_2017,reichstein_deep_2019}. Purely data-driven examples include Sentinel-1 SM retrieval \cite{zhu_cross-resolution_2024} and the FLUXCOM-X \cite{nelson_x-base_2024} upscaling of evapotranspiration (ET), gross primary productivity (GPP) and net ecosystem exchange (NEE). Physics-informed and knowledge-guided approaches have also considered constrained SM prediction \cite{wang_deep_2023,yu_field-scale_2025}, and combined process-model knowledge, remote sensing and ML for carbon-cycle quantification \cite{liu_knowledge-guided_2024}. However, trained ML and DL models remain dependent on the statistical relationships and environmental conditions represented by their training data. Ecohydrological states and fluxes depend on temporal memory, physical constraints and non-stationary responses to climate and extremes, making model performance sensitive to shifts in sensor, region, ecosystem, scale, hydroclimate and event magnitude \cite{ma_transfer_2024,reichstein_deep_2019,shen_differentiable_2023}. Predictions from purely data-driven models quantify statistical associations; process attribution and causal explanation require independent process evidence \cite{reichstein_deep_2019,shen_differentiable_2023}. Broader geographic coverage and transfer learning can reduce labelling requirements \cite{ma_transfer_2024}. Out-of-domain reliability nevertheless remains limited. Global datasets can still leave environmental combinations outside the area of applicability (AOA) \cite{meyer_predicting_2021,meyer_machine_2022}; documented limitations include limited AOA coverage in the SM mapping domain \cite{yu_spatial_2025}, unacceptable transfer performance in mountainous terrain \cite{zhu_cross-resolution_2024}, and poor geographic generalisation, with a randomly initialised model remaining competitive when labelled data were sufficient \cite{gordon_mmearth-bench_2026}. These limitations reveal a need for globally transferable EO representations that downstream ecohydrological models can use across sensors and regions \cite{lu_vision_2025,xiao_foundation_2025}.

An Earth observation foundation model (EOFM) is a large pretrained model that learns reusable representations from extensive, heterogeneous EO data and adapts them to downstream tasks through linear probing, fine-tuning, parameter-efficient adapters or prompting \cite{lu_vision_2025,xiao_foundation_2025}. The design of EOFM accounts for the sensor-specific observation physics, geolocation, spatial scale, temporal dynamics and the multimodal structure of EO data \cite{xiao_foundation_2025,zhu_foundations_2026}. Existing models already support multitemporal and multispectral learning, radar--optical alignment, wavelength-conditioned processing across sensors and generative multimodal pretraining \cite{cong_satmae_2022,fuller_croma_2023,xiong_neural_2025,jakubik_terramind_2025,szwarcman_prithvi-eo-20_2026}. These capabilities are relevant to ecohydrology because complementary sensors provide partial information about coupled water, energy, and carbon dynamics in vegetation and soil. A growing body of reviews and surveys has provided valuable field-wide syntheses of model architectures, pretraining strategies, EO modalities, benchmarks and general downstream applications \cite{jiao_brain-inspired_2023,huo_when_2025,lu_vision_2025,xiao_foundation_2025,zhang_when_2025,zhou_advances_2025,hong_foundation_2026}. Table~\ref{tab:eofm_review_scope} summarises the complementary scopes of representative studies. Within this survey landscape, we found no existing synthesis that jointly evaluates EOFM representations against ecohydrological observability, process timescales, reference independence. Such a synthesis must consider how seasonal trajectories, event onset and recovery, and antecedent conditions encode process memory; how scale-dependent and model-derived reference data affect evaluation; and how predictions preserve physical bounds and water--energy--carbon coupling under geographic, climatic, temporal and sensor shifts \cite{reichstein_deep_2019,shen_differentiable_2023,zhu_foundations_2026}. EOFMs also inherit the sampling biases, measurement uncertainties and processing assumptions of their pretraining and adaptation data. These requirements motivate an ecohydrology-specific assessment of when reusable EO representations provide scientifically meaningful information about dynamic states and fluxes.

These gaps lead to an overarching question: under what conditions can an EOFM provide reliable and scientifically interpretable information about coupled water, energy and carbon dynamics in vegetation and soil across space and time? Addressing this question requires an integrated assessment of how ecohydrological information is inferred, what EOFMs can represent, what their applications have demonstrated and how the resulting claims should be evaluated. We therefore divide the overarching question into four more specific research questions (RQs):
\begin{enumerate}
  \item How do EO signals and complementary data support inference of ecohydrological properties, states, fluxes, mechanisms and decision-relevant outcomes, and where do observation physics, reference uncertainty, scale and process constraints enter this inference chain (Section~\ref{sec:ecohydrol_inference})?
  \item Which EO modalities, spatial and temporal supports, representation strategies and adaptation routes are covered by current EOFMs, and which capabilities required for ecohydrology remain underrepresented (Section~\ref{sec:eofm_meta})?
  \item What EOFM capabilities have been demonstrated across ecohydrological target classes, and how does the strength of this evidence depend on target observability, reference independence and evaluation under environmental or sensor shifts (Section~\ref{sec:eofm_capabilities})?
  \item Which benchmark designs and diagnostic criteria are needed to evaluate accuracy, temporal dynamics, physical consistency, uncertainty, transferability and readiness for scientific or operational use (Section~\ref{sec:benchmark})?
\end{enumerate}

Together, the four RQs define the inference, establish what current EOFMs can represent, assess what their applications have demonstrated, and determine how those claims should be evaluated. The review follows this sequence: the ``\hyperref[sec:ecohydrol_inference]{Inference of Ecohydrology}'' section establishes the observation-to-inference hierarchy and its uncertainty pathways; the ``\hyperref[sec:eofm_meta]{Meta-Analysis of EOFMs}'' section maps the EOFM design space; the ``\hyperref[sec:eofm_capabilities]{Capabilities of EOFMs across Ecohydrological Tasks}'' section synthesises evidence across ecohydrological target classes; and the ``\hyperref[sec:benchmark]{Benchmarking EOFMs for Ecohydrology}'' section examines existing benchmarks and develops a target-first evaluation framework. The ``\hyperref[sec:discussion]{Discussion}'' section integrates the findings and scientific boundaries across the four questions, the ``\hyperref[sec:prospects]{Prospects}'' section sets out priorities for trustworthy and process-aware representations, and the ``\hyperref[sec:conclusion]{Conclusion}'' section concludes the review.

\section{Inference of Ecohydrology}
\label{sec:ecohydrol_inference}

\subsection{A Hierarchical Framework}
\label{sec:hierarchical_framework}

The inference of ecohydrology is the process of combining measurements with physical relationships, models and contextual information to estimate states, fluxes, mechanisms or decision-relevant conditions at a specified spatial and temporal support \cite{jiao_remote_2026,mohanty_scale_2026,eller_scaling_2026}. Remote sensing provides spatially and temporally extensive data for quantifying plant--soil--atmosphere interactions and connecting hydrological, ecological and atmospheric processes \cite{jiao_remote_2026}. Inference is intrinsic to EO-based ecohydrology because sensors record radiance, brightness temperature or backscatter, whereas ecohydrological questions commonly concern quantities such as SM, vegetation water status, ET, GPP or plant--water coupling that are retrieved or modelled from those signals \cite{peng_roadmap_2021,fisher_evapotranspiration_2026}. Inference also arises when information is transferred across observational support because point measurements, flux-tower footprints and EO pixels sample different portions of heterogeneous landscapes, and scaling local process understanding to regional or global estimates requires assumptions about aggregation, representativeness and nonlinear process responses \cite{mohanty_scale_2026,crow_upscaling_2012}.

\begin{figure*}[]
\centering
\includegraphics[width=1.\textwidth]{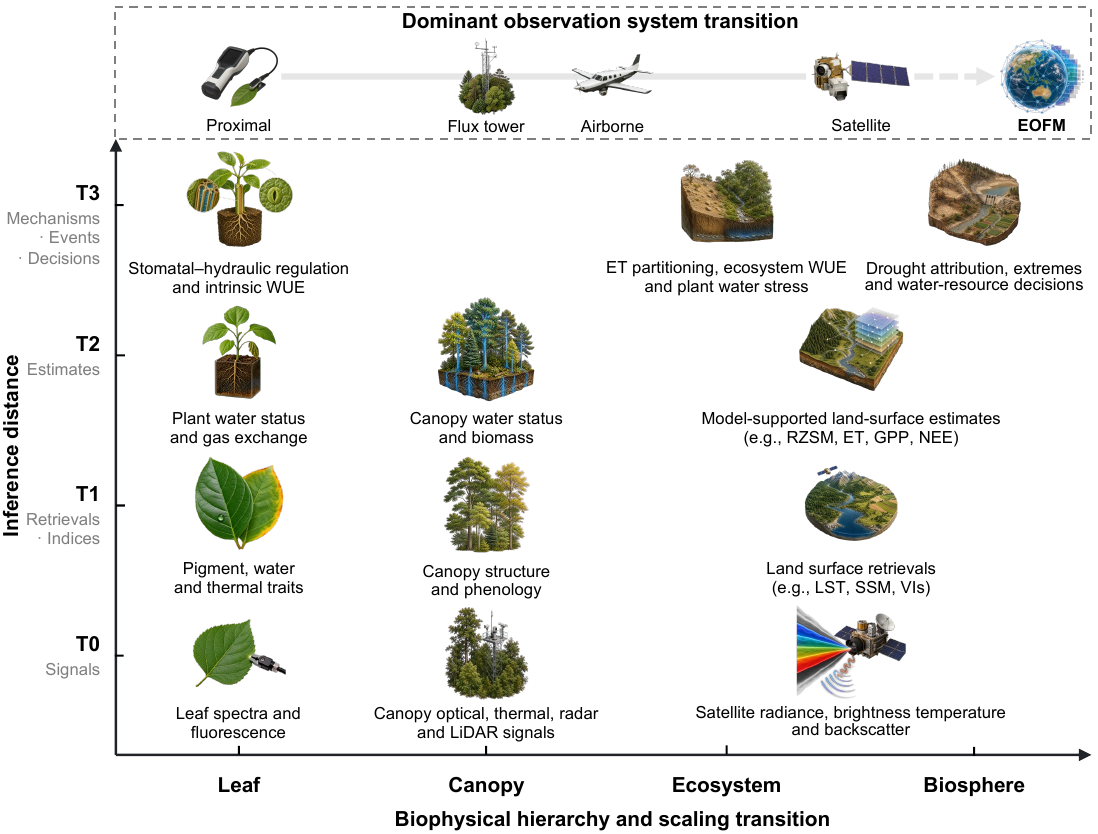}
\caption{A hierarchical framework linking biophysical organisation, inference distance and dominant observation systems in ecohydrology. The horizontal progression follows the scaling of coupled processes from leaves, through canopies and ecosystems, to the biosphere. The vertical tiers distinguish calibrated signals (T0), retrievals and indices (T1), estimated states and fluxes (T2), and mechanisms, events and decisions (T3); examples show representative quantities at different combinations of scale and inference distance. The upper band shows the transition from proximal sensors and flux towers to airborne and satellite EO, which could ultimately be embedded into EOFMs. Moving among tiers and scales introduces pathway-specific uncertainty from sensing, retrieval, forcing, partitioning, upscaling and model assumptions.}
\label{fig:01_ecohydrol_hierarchy}
\end{figure*}

Drawing on these complementary EO, scaling and plant-hydraulic perspectives, Fig.~\ref{fig:01_ecohydrol_hierarchy} synthesises two linked hierarchies: biophysical organisation from the leaf scale to the biosphere, and inference from measured signals through retrievals and estimated states or fluxes to mechanisms, events and decisions \cite{jiao_remote_2026,mohanty_scale_2026,eller_scaling_2026}. Within the biophysical hierarchy, proximal measurements resolve leaf- and canopy-scale physiology, tower and airborne systems integrate responses over stands and ecosystems, and repeated satellite acquisitions extend coverage towards regional and global scales \cite{jiao_remote_2026,mohanty_scale_2026,eller_scaling_2026}. This progression increases spatial coverage, temporal repetition and data volume, while larger observational support aggregates heterogeneity in vegetation, soil, topography, climate and management, increasing the importance of scale matching and representativeness \cite{crow_upscaling_2012,yu_empirical_2024,peng_spatial_2025,mohanty_scale_2026}. Across both hierarchies, ecohydrological quantities emerge from coordinated processes among biological and environmental compartments, and sensing, modelling and scale transfer create pathway-specific uncertainty between the recorded signal and its scientific interpretation \cite{jiao_remote_2026,mohanty_scale_2026,eller_scaling_2026}. The inference hierarchy is organised into four tiers:

\begin{itemize}
  \item \textit{Tier~0: calibrated signals.} Leaf and canopy spectra, fluorescence-containing radiance, thermal radiance, microwave brightness temperature and radar backscatter remain closest to sensor measurement. Their interpretation already depends on calibration, viewing geometry, atmospheric effects and radiative transfer models \cite{jiao_remote_2026,mohammed_remote_2019}.

  \item \textit{Tier~1: retrievals and indices.} Sensor signals are transformed into vegetation indices (VIs), leaf area index (LAI), phenological metrics, land surface temperature (LST), surface SM and retrieved solar-induced chlorophyll fluorescence (SIF). These quantities depend on empirical or radiative-transfer inversion and remain sensitive to canopy structure, background conditions, emissivity, roughness, vegetation attenuation and atmospheric correction \cite{zeng_optical_2022,jiang_inconsistencies_2017,babaeian_ground_2019,mohammed_remote_2019}.

  \item \textit{Tier~2: estimated states and fluxes.} Retrievals are combined with meteorological forcing, spatial transfer and model structure to estimate plant or canopy water status, biomass, root-zone SM, ET, GPP and NEE \cite{jiao_remote_2026,fatichi_modeling_2016}. SM pathways use optical, thermal, passive-microwave or active-microwave signals with different sensing depths and vegetation sensitivities; vapour pressure deficit (VPD) can be derived from reanalysis, thermal-infrared or passive-microwave temperature and humidity fields; and GPP and ET products use light-use-efficiency, process-based or data-driven models \cite{wang_satellite_2009,peng_roadmap_2021,du_global_2018,running_continuous_2004,jung_fluxcom_2019,zhang_review_2016,chen_evolution_2020}. Optical absorption features and microwave vegetation attenuation likewise constrain complementary aspects of plant water content \cite{konings_macro_2019,jiao_remote_2026}.

  \item \textit{Tier~3: mechanisms, events and decisions.} Multiple states and fluxes are interpreted jointly to assess stomatal--hydraulic regulation and intrinsic WUE, partition ET among transpiration, soil evaporation and canopy interception, quantify ecosystem WUE or biomass production, attribute drought and compound extremes, or inform water-resource decisions \cite{eller_scaling_2026,kool_review_2014,cai_remote_2021,jiao_remote_2026}. These claims require temporal context and evidence across several variables because water availability, atmospheric demand and carbon uptake can interact nonlinearly \cite{novick_increasing_2016,grossiord_plant_2020}.
\end{itemize}

The progression represents increasing inference distance from measurement and increasing dependence on auxiliary data, model assumptions and independent process evidence. Successive scientific interpretations of EO are connected by pathways along which uncertainty accumulates differently because every transformation from calibrated signals to retrievals, estimated states and fluxes, and ultimately mechanisms or decisions inherits preceding errors and introduces new assumptions. For example, fluorescence-containing radiance can be transformed into retrieved SIF, an SIF-constrained estimate of photosynthetic carbon uptake, and an interpretation of vegetation stress or process response; this progression adds uncertainty from calibration and radiative transfer, retrieval inversion, meteorological forcing and modelling, and mechanistic interpretation \cite{mohammed_remote_2019,novick_increasing_2016,grossiord_plant_2020}. Comparable dependencies arise from atmospheric correction, emissivity and surface roughness during signal processing and retrieval, and from parameterisation, gap filling, flux partitioning and model structure when states and fluxes are estimated \cite{jiao_remote_2026,fatichi_modeling_2016}. Reference data introduce further scale- and process-dependent uncertainty: satellite SM and point measurements differ in sensing depth and spatial support \cite{crow_upscaling_2012,gruber_validation_2020}; tower-based GPP depends on partitioning NEE into photosynthesis and respiration \cite{reichstein_separation_2005}; and tower-based ET evaluation is influenced by flux footprints, gap filling and incomplete energy-balance closure \cite{jung_fluxcom_2019,fisher_evapotranspiration_2026}. Temporal support adds another constraint because phenology, SM dry-down, drought onset and recovery, irrigation pulses and carbon--water coupling depend on sequences that resolve antecedent conditions, timing, lags and nonlinear interactions between soil water supply and atmospheric demand \cite{novick_increasing_2016,grossiord_plant_2020}.
Agreement with a retrieved or model-derived products therefore establishes fidelity to the selected product. Independent recovery of the underlying state, flux or mechanism requires reference evidence matched to the intended interpretation, biophysical support and temporal dynamics \cite{jiao_remote_2026,mohanty_scale_2026,fatichi_modeling_2016}. Together, these requirements frame the role of EOFMs in ecohydrology by specifying the scale-aware, temporally resolved and process-relevant information their representations must retain for defensible inference.

\subsection{EOFM for Ecohydrology}
\label{sec:eofm4ecohydrol}

\begin{figure*}[!b]
\centering
\includegraphics[width=1.0\textwidth]{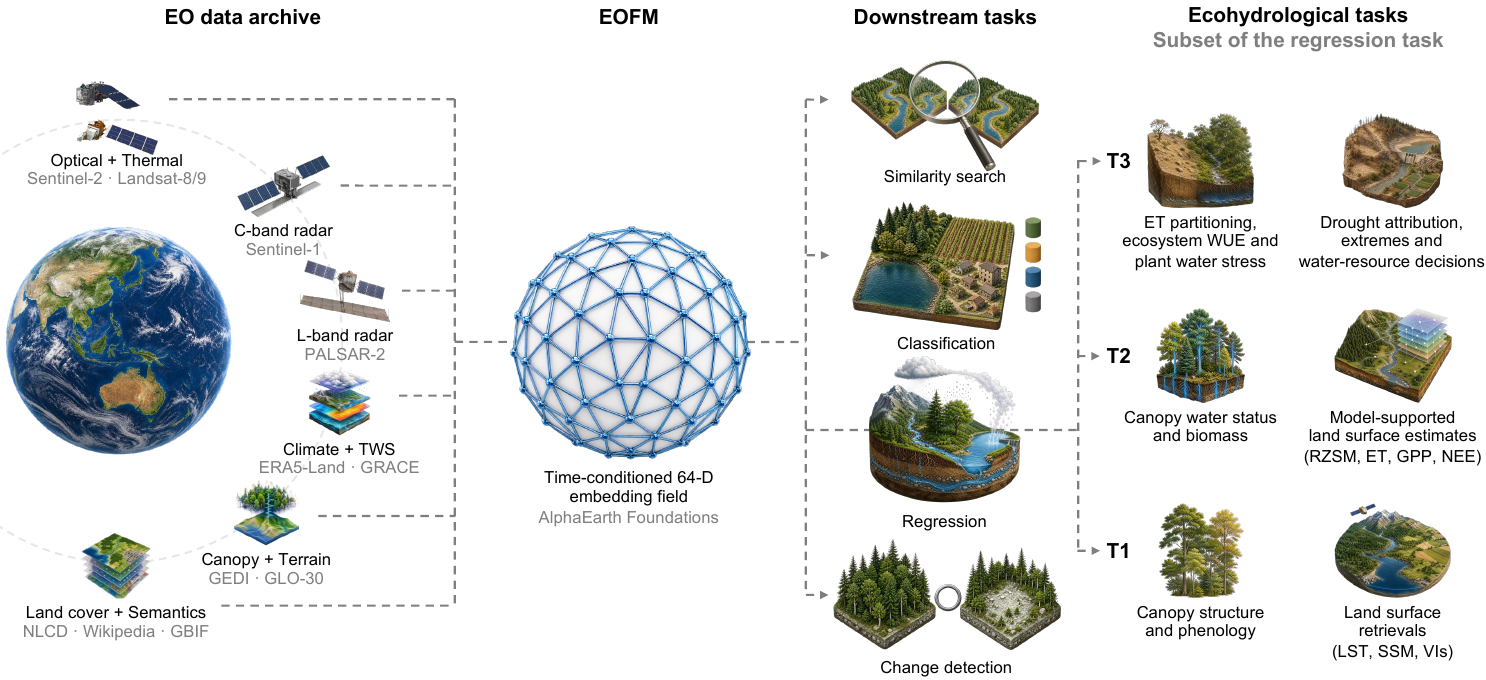}
\caption{Illustrative pathway from multisource EO pretraining to ecohydrological inference, using AlphaEarth Foundations as an example. During training, optical and thermal imagery, C- and L-band radar, climate and terrestrial-water-storage data, canopy and terrain measurements, and land-cover and semantic information serve as model inputs or prediction targets for learning a time-conditioned 64-dimensional embedding field; only the designated input sources are required to generate the embedding field at inference \cite{brown_alphaearth_2025}. The general downstream operations comprise similarity search, classification, regression and change detection. The ecohydrological examples expand the regression branch according to Fig.~\ref{fig:01_ecohydrol_hierarchy}: T1 represents retrievals and indices, T2 estimated states and fluxes, and T3 mechanisms, events and decisions.}
\label{fig:02_eofm4ecohydrol}
\end{figure*}

Ecohydrology and EOFMs meet through representation learning: EO archives repeatedly sample the spectral, spatial and temporal expressions of coupled water, energy, vegetation and carbon dynamics, while EOFM pretraining seeks statistical structure that can be transferred before an ecohydrological target is introduced. Masked reconstruction, cross-modal prediction and contrastive alignment provide supervisory signals from the data themselves by requiring a model to recover withheld spatial, spectral or temporal content and to align co-located observations from complementary sensors \cite{cong_satmae_2022,fuller_croma_2023,nedungadi_mmearth_2025}. Pretraining across locations, seasons and sensors can consequently encode regularities associated with vegetation phenology, surface wetness, thermal conditions, canopy structure and landscape context; downstream adaptation then links these representations to sparsely measured states, properties or events \cite{feng_tessera_2026,nedungadi_mmearth_2025}. Fig.~\ref{fig:02_eofm4ecohydrol} illustrates this pathway using AlphaEarth Foundations as an example, which reconciles spatially and temporally irregular multisource data as model inputs or prediction targets to learn a time-conditioned 64-dimensional embedding field for downstream tasks when labels are sparse \cite{brown_alphaearth_2025}. Emerging applications provide initial evidence that these embeddings can support ecohydrological prediction when combined with dynamic observations and forcing: they improved station-held-out LST recovery when fused with GOES-18 observations and auxiliary predictors \cite{lee-burkhart_geospatial_2026}, supported 10 m daily urban ET mapping when combined with Sentinel-2 vegetation information and meteorological forcing \cite{jiang_transformer-based_2026}, and improved spatial out-of-sample streamflow prediction when used as basin descriptors across 455 Australian catchments \cite{ou_foundation-scale_2026}. These hybrid results demonstrate transferable environmental context within AlphaEarth embeddings and motivate closer examination of the spatial, temporal and adaptation conditions under which that context supports ecohydrological inference.

Most existing EOFMs have been designed to learn reusable representations of multiple EO streams through radar--optical alignment, multimodal prediction targets, wavelength-conditioned transfer across sensors, generative translation among data sources and assimilation of heterogeneous measurement contexts \cite{fuller_croma_2023,xiong_neural_2025,nedungadi_mmearth_2025,jakubik_terramind_2025,brown_alphaearth_2025}.
For ecohydrological applications, the relevance of these representations depends on how effectively they encode spatial organisation and temporal evolution: spatial information links local signals to terrain, vegetation structure and hydrological connectivity across nested supports, while temporal information retains seasonal trajectories, antecedent conditions and lagged responses across event-to-interannual timescales \cite{jiao_remote_2026,peng_roadmap_2021,mohammed_remote_2019,babaeian_ground_2019,fisher_evapotranspiration_2026}.
Multimodal capability concerns the information that is jointly learned and made accessible to downstream inference. Input catalogues identify model exposure; a capability assessment must further determine whether the representation reconciles asynchronous and scale-mismatched sources, preserves the spatial context and temporal memory required by the target process, makes the relevant information available at inference, and supports adaptation through frozen embeddings, linear probes, parameter-efficient methods or full fine-tuning. These distinctions are consequential. Global embedding coverage does not establish uniform modality, acquisition-time or data-quality support, and estimating dynamic ecohydrological states and fluxes may still require contemporary EO and meteorological forcing \cite{brown_alphaearth_2025,lee-burkhart_geospatial_2026,jiang_transformer-based_2026}. A thorough synthesis is then needed to document model inputs and prediction targets, observation pathways, spatial and temporal support, pretraining objectives, downstream interfaces, adaptation routes and the evidence used to evaluate ecohydrological relevance. Existing reviews describe architectures, modalities and general EO applications \cite{lu_vision_2025,xiao_foundation_2025,hong_foundation_2026}. An ecohydrology-specific roadmap connecting these characteristics to observation physics, process timescales and evidentiary requirements remains undeveloped. We therefore collate and critically synthesise this design space, identifying the multimodal capabilities implemented across EOFMs and the extent to which their ecohydrological relevance has been demonstrated.

\section{Meta-Analysis of EOFMs}
\label{sec:eofm_meta}

\begin{figure}[t]
\centering
\includegraphics[width=0.50\textwidth]{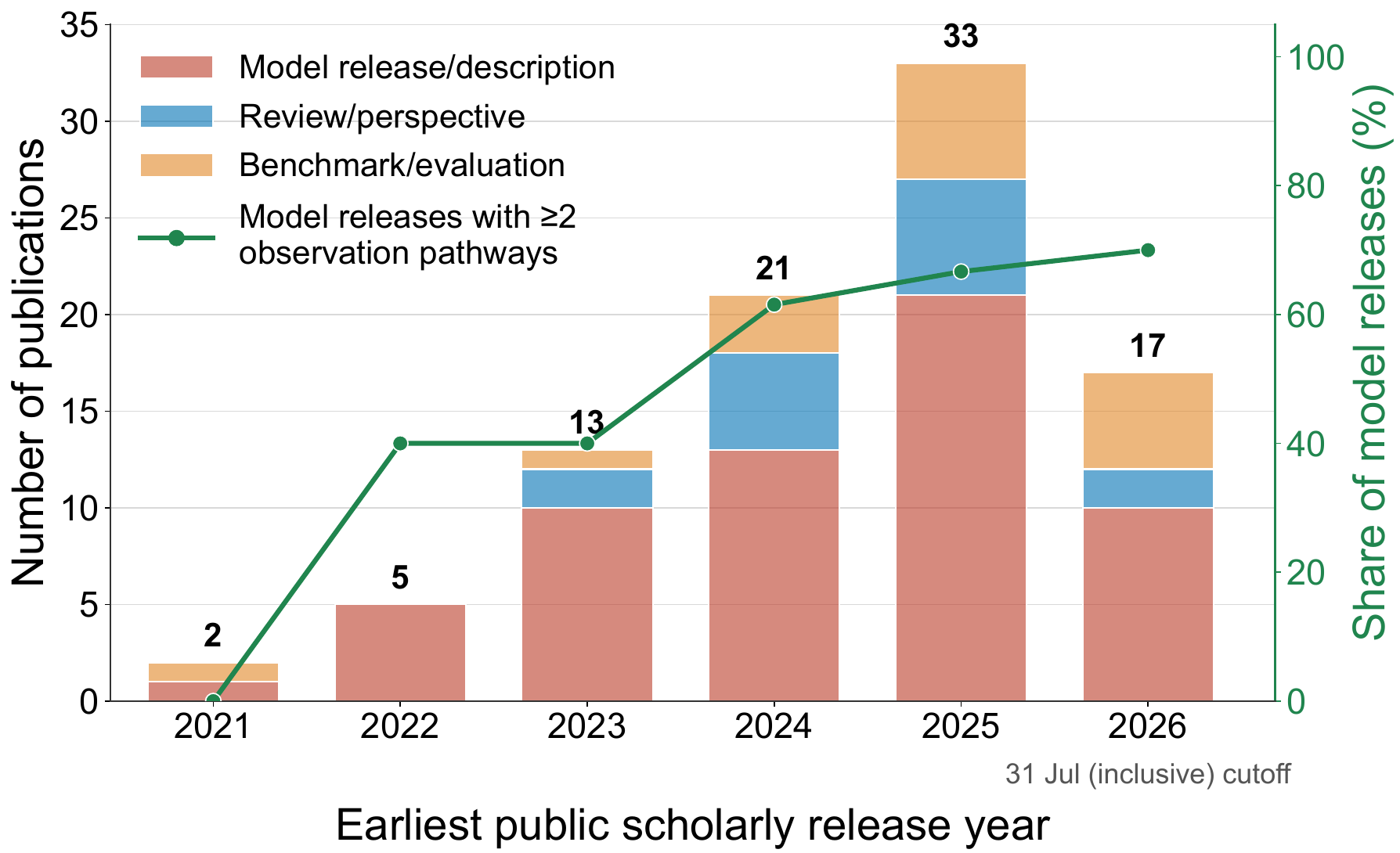}
\caption{Publication growth and diversification in EOFM research. Annual counts are assigned to the year of earliest public scholarly release and stacked by publication role. The green line indicates the percentage of model releases using at least two coded EO data pathways; reviews/perspectives and benchmarks/evaluations are excluded from this denominator. The 2026 record is partial and includes releases available through 31 July.}
\label{fig:03_publication_trends}
\end{figure}

EOFM research has expanded rapidly since 2021, accompanied by a growing body of review and evaluation studies and an increasing share of model releases drawing on multiple EO data pathways (Fig.~\ref{fig:03_publication_trends}). We characterised this development through a meta-analysis of publications available by 31 July 2026. The EOFM literature remains emerging with many contributions disseminated through conference proceedings, preprints and model documentation. The literature search therefore covered \emph{OpenAlex}, \emph{arXiv}, \emph{Semantic Scholar} and \emph{DBLP}, combining the keyword ``foundation model'' with ``remote sensing'', ``Earth observation'', ``geospatial'', ``satellite'', ``multimodal'' and ``temporal''. The search returned 2,106 records; linkage by DOI, \emph{arXiv} identifier and normalised title, followed by relevance screening, yielded 722 deduplicated candidates whose eligibility and technical attributes were verified against primary articles, official proceedings, repositories and model documentation. Finally, by removing the deduplications and reconciling model identity, the meta-analysis contains 60 model/version records representing 52 families; the broader publication-trend corpus comprises 60 model descriptions, 15 reviews or perspectives and 16 benchmark or evaluation studies. Forecast-focused weather and Earth-system FMs, generic data resources, specialised retrieval models without reusable representations and records with unresolved model identities were excluded. Each version was coded independently for input pathways, representation capabilities, adaptation modes and downstream outputs; family identifiers were retained for sensitivity analysis, and unreported characteristics remained missing. The resulting evidence base supports an assessment of EOFM growth, multimodal composition and spatiotemporal design.

\subsection{Overall Trends in EOFM Development}
\label{sec:eofm_trends}

The development of EOFMs is summarised through three related characteristics.

\subsubsection{Publication Growth}
Fig.~\ref{fig:03_publication_trends} shows rapid expansion in model development and critical evaluation. Model releases increased from one in 2021 to 21 in 2025, and the broader annual publication count rose from two to 33. The partial 2026 record already contains ten model releases and 17 total publications through 31 July. The growing contributions of reviews and benchmarks since 2023 show that new model development is increasingly accompanied by synthesis, comparison and independent evaluation \cite{lu_vision_2025,marsocci_pangaea_2026,li_reobench_2026}.

\subsubsection{Multimodal Development}
The share of model releases using at least two EO data pathways increased from 0\% in 2021 to 66.7\% in 2025 and 70.0\% in the partial 2026 record (Fig.~\ref{fig:03_publication_trends}). Early-year percentages are based on small numbers of releases and require cautious interpretation. The pathway groups in Fig.~\ref{fig:04_eofm_map} further show the increasing presence of optical--microwave and broader multisensor or ancillary configurations after 2023. These counts describe input breadth; cross-modal alignment, missing-modality robustness and process-specific complementarity require separate evaluation.

\subsubsection{Spatiotemporal Diversity}
\begin{figure*}[t]
\centering
\includegraphics[width=1.\textwidth]{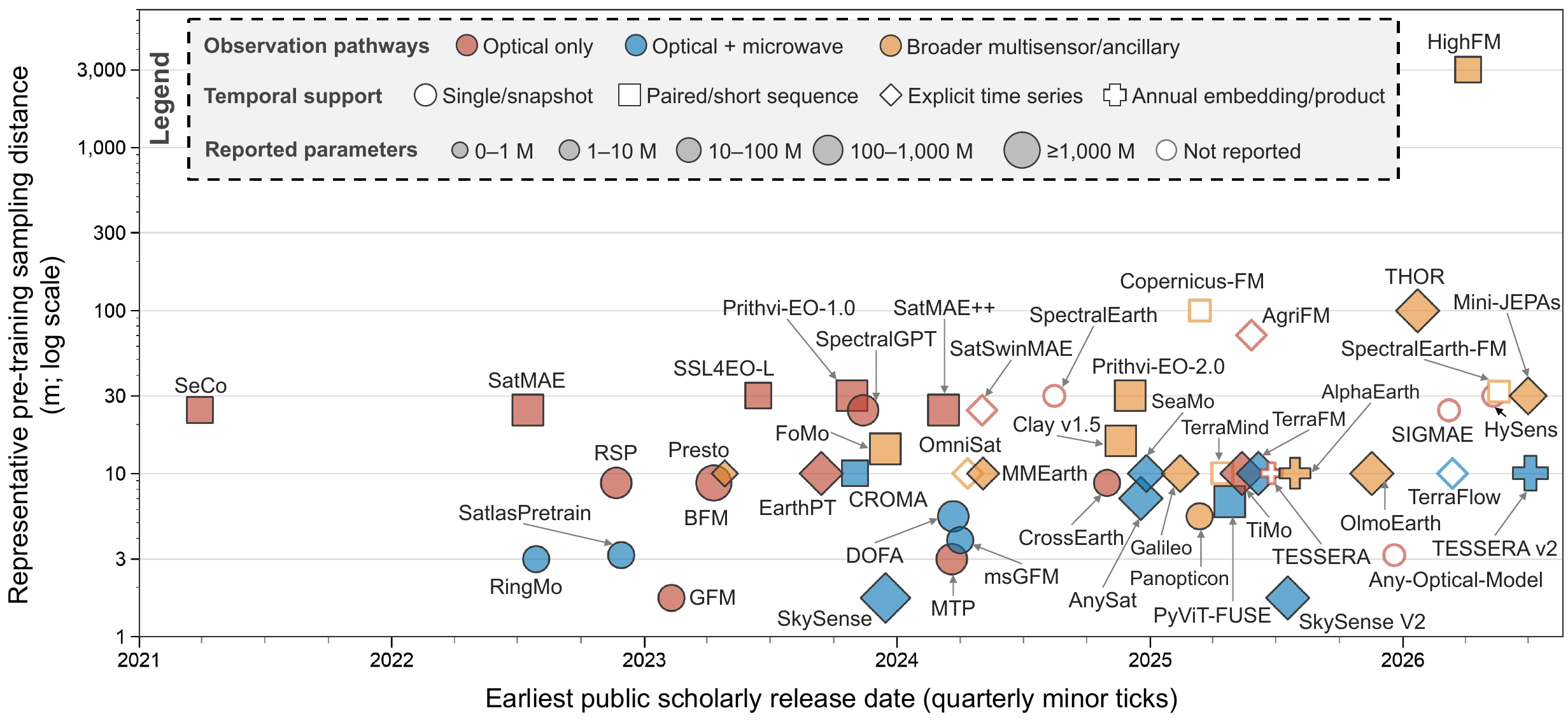}
\caption{Sampling distance and temporal design of shortlisted EOFMs. The 49 plotted model/version releases are positioned by earliest public scholarly release month and one pre-training sampling-distance value. A single reported ground sampling distance is used directly; when only a numeric interval was available, its geometric midpoint provides a transparent plotting proxy rather than an empirical median. Colour identifies the retained observation-pathway group, marker shape identifies temporal support, and marker area represents reported parameters using a base-10 logarithmic transformation.}
\label{fig:04_eofm_map}
\end{figure*}

Fig.~\ref{fig:04_eofm_map} maps 49 releases by pre-training sampling distance, temporal support, EO pathway group and reported parameter count. The widening distribution since 2023 extends from fine-resolution landscape information to kilometre-scale geostationary data, together with designs ranging from single snapshots to explicit time series and annual embedding products. Sampling distance, temporal category and parameter count describe model design; their ecohydrological relevance depends on evidence matched to the spatial support, response time and target process.

\onecolumn

\begin{multicols}{2}
\subsection{Quantitative Analysis of EOFMs}
\label{sec:eofm_evidence}

We quantitatively analysed the 60 eligible EOFM releases based on their original release papers and reports. The applications documented in this section comprise the downstream evaluations reported alongside the original model releases. We assessed 4 design and validation dimensions: input pathways; spatiotemporal and model scale; adaptation and outputs; and ecohydrological relevance. Table~\ref{tab:eofm_ecohydrol} compares model design, task transfer and validation directness across the 60 releases; herein we also summarise patterns within each dimension.

\subsubsection{Input Pathways}
Reflected optical data occur in 57 releases (95.0\%), microwave data in 34 releases (56.7\%; predominantly active SAR), ancillary, retrieved or modelled inputs in 15 releases (25.0\%) and thermal data in 5 releases (8.3\%). The eligible corpus contains 0 releases with an explicit passive-microwave-emission or SIF pre-training pathway. The coding assigns hyperspectral reflectance to the optical pathway and reserves the fluorescence pathway for explicit SIF data. Multimodal inputs occur in 34 releases (56.7\%). Current multimodal development remains concentrated in optical--SAR fusion and contextual layers, with sparse representation of distinct signal-generation pathways relevant to ecohydrological processes.

\subsubsection{Spatiotemporal and Model Scale}
Numerical pre-training resolution ranges are reported for 51 releases (85.0\%); among these releases, 49 satisfy the retained pathway criteria in Fig.~\ref{fig:04_eofm_map}. Their representative sampling distances range from approximately 1.7~m to 3~km, with a median of 10~m and an interquartile range of 8.7--24.5~m. The distribution of temporal support comprises 21 single images or snapshots (35.0\%), 16 paired observations or short sequences (26.7\%), 20 explicit time series (33.3\%) and 3 annual embedding products (5.0\%). These categories describe the observations supplied during pre-training; process memory requires separate evaluation of cadence, sequence length, event timing and lagged responses. Parameter counts are reported for 43 releases (71.7\%), with a median of 300 million (interquartile range: 88--600 million) and a range from 0.402 million to 14.7 billion. Sampling distance and parameter count describe input and model scale. Process support and ecohydrological skill require empirical evaluation.

\subsubsection{Adaptation and Outputs}
Adaptation routes overlap across releases. Fine-tuning is reported for 55 releases (91.7\%), linear probing for 20 releases (33.3\%), frozen-encoder or frozen-embedding use for 15 releases (25.0\%), parameter-efficient fine-tuning for 3 releases (5.0\%) and zero-shot evaluation for 2 releases (3.3\%). Downstream evaluations concentrate on segmentation (53 releases; 88.3\%) and classification (51 releases; 85.0\%), followed by regression (19 releases; 31.7\%), change detection (18 releases; 30.0\%) and object detection (15 releases; 25.0\%); forecasting is reported for 1 release. Current evaluations therefore centre on task-specific semantic mapping, with substantially fewer tests of dynamic continuous prediction, parameter-efficient adaptation and zero-shot transfer.

\subsubsection{Ecohydrological Relevance}

Direct ecohydrological evaluation is reported for 6 releases (10.0\%). Four releases (6.7\%) use independent field, \emph{in situ}, or flux-reference data. HySens evaluates surface SM, SOC and water chlorophyll against field measurements; Presto evaluates LFMC against 1,578 field samples using a geographically separated test; Prithvi-EO-2.0 evaluates GPP partitioned from eddy-covariance NEE at 37 sites using leave-one-year-out validation; and StefaLand evaluates \emph{in situ} SM and observed streamflow under spatial holdouts \cite{zhao_hysens_2026,tseng_lightweight_2024,szwarcman_prithvi-eo-20_2026,kraabel_stefaland_2026}. Two releases (3.3\%) use product or proxy references: AlphaEarth Foundations evaluates monthly OpenET ET regression, and Mini-JEPAs evaluate SMAP SM, aridity and precipitation products \cite{brown_alphaearth_2025,rahman_sensor-specialized_2026}. One release, Prithvi-EO-2.0, provides flux-tower-referenced evaluation of a partitioned carbon flux. The eligible corpus contains no independent flux-tower validation of ET, and no release evaluates NEE directly. The other 54 releases (90.0\%) establish indirect relevance through classification, segmentation, change detection, retrieval or related land-surface tasks. Reference variables derived from retrievals, reanalysis or models carry their source assumptions, smoothing, spatial and temporal support, and uncertainty into downstream evaluation.
\end{multicols}

\small
\setlength{\tabcolsep}{0.8pt}
\renewcommand{\arraystretch}{1.00}
\setlength{\defaultaddspace}{1pt}
\setlength{\LTleft}{0pt}
\setlength{\LTright}{0pt}
\setlength{\LTcapwidth}{\textwidth}
\begin{longtable}{@{}>{\raggedright\arraybackslash}p{0.078\textwidth}>{\raggedright\arraybackslash}p{0.100\textwidth}>{\raggedright\arraybackslash}p{0.140\textwidth}>{\raggedright\arraybackslash}p{0.202\textwidth}>{\raggedright\arraybackslash}p{0.078\textwidth}>{\centering\arraybackslash}p{0.090\textwidth}>{\raggedright\arraybackslash}p{0.282\textwidth}@{}}
\caption{Comparison of the 60 eligible EOFM model/version releases by model design, task transfer and directness of reported ecohydrological relevance. Rows are ordered by the month of earliest verified public release and then alphabetically by model name within each month.}\label{tab:eofm_ecohydrol}\\
\toprule
\textbf{First release}\textsuperscript{a} & \textbf{Model name} & \textbf{Backbone and reported parameters}\textsuperscript{b} & \textbf{Pre-training inputs and pathway}\textsuperscript{c} & \textbf{Spatial resolution (m)}\textsuperscript{d} & \textbf{Temporal support}\textsuperscript{e} & \textbf{Ecohydrological relevance}\textsuperscript{f}, \textbf{adaptation}\textsuperscript{g}, \textbf{outputs}\textsuperscript{h} \\
\midrule
\endfirsthead
\caption[]{Comparison of EOFM model/version releases by model design, task transfer and directness of reported ecohydrological relevance (continued).}\\
\toprule
\textbf{First release}\textsuperscript{a} & \textbf{Model name} & \textbf{Backbone and reported parameters}\textsuperscript{b} & \textbf{Pre-training inputs and pathway}\textsuperscript{c} & \textbf{Spatial resolution (m)}\textsuperscript{d} & \textbf{Temporal support}\textsuperscript{e} & \textbf{Ecohydrological relevance}\textsuperscript{f}, \textbf{adaptation}\textsuperscript{g}, \textbf{outputs}\textsuperscript{h} \\
\midrule
\endhead
\midrule
\multicolumn{7}{r}{\small\emph{Continued on next page}}\\
\endfoot
\bottomrule
\multicolumn{7}{@{}p{\dimexpr\textwidth-4\tabcolsep\relax}@{}}{\small
\textsuperscript{a} The earliest verified public release date is shown as an abbreviated month and year. Rows sharing a release month are ordered alphabetically by model name.\par
\textsuperscript{b} The parameter count refers to the reported encoder or operational representation model. M = million parameters; B = billion parameters; n/r = not reported for the cited model/version.\par
\textsuperscript{c} Pre-training pathway codes cover data used as model inputs or pre-training prediction targets. O = reflected optical data only; M = microwave data, including synthetic aperture radar (SAR), used alone or added to the O pathway; B = broader multisensor or ancillary data, including thermal or contextual inputs.\par
\textsuperscript{d} Spatial resolution is the native or analysis-ready resolution, in metres (m), of the principal pre-training inputs. A range indicates that the inputs span multiple spatial resolutions.\par
\textsuperscript{e} Temporal support describes pre-training sampling: SN = single image or snapshot; PS = paired observations or a short sequence; TS = explicit time series; AP = annual embedding or product. These category codes refer to pre-training support. Downstream tasks may use different cadences or temporal windows.\par
\textsuperscript{f} Ecohydrological relevance: D-I = a direct state or flux target evaluated against independent field, \emph{in situ} or flux reference data; D-P = a direct target evaluated against a product or proxy; I = indirect relevance established through related tasks without direct state or flux validation.\par
\textsuperscript{g} Adaptation: FE = frozen embeddings or a frozen encoder; LP = linear probing; FT = fine-tuning; PEFT = parameter-efficient fine-tuning; ZS = zero-shot evaluation.\par
\textsuperscript{h} Outputs: Cls. = classification; Seg. = segmentation; Det. = object detection; CD = change detection; Reg. = regression; Fcst. = forecasting; Gen. = generation; Ret. = retrieval; SR = super-resolution.}\\
\endlastfoot

\mbox{Mar. 2021} & SeCo \cite{manas_seasonal_2021} & ResNet-50; 23.5 M. & 200,000 global Sentinel-2 seasonal views. \textbf{[O]} & 10--60 & PS & LP/FT $\rightarrow$ Cls., Seg. \textbf{I}: Seasonal vegetation and change context; no direct state or flux test. \\
\addlinespace[2pt]
\mbox{Jul. 2022} & RingMo \cite{sun_ringmo_2023} & Swin/ViT; 87.8 M. & Two million optical and SAR images. \textbf{[M]} & 0.3--30 & SN & FT $\rightarrow$ Cls., Det., Seg., CD. \textbf{I}: Optical--SAR supports flood and surface mapping; no continuous-state or flux test. \\
\mbox{Jul. 2022} & SatMAE \cite{cong_satmae_2022} & ViT-L; 307 M. & fMoW-Sentinel 13-band imagery and short temporal stacks. \textbf{[O]} & 10--60 & PS & LP/FT $\rightarrow$ Cls., Seg. \textbf{I}: Crop, phenology and flood context; short sequences do not establish process memory. \\
\mbox{Nov. 2022} & RSP \cite{wang_advancing_2023} & ViT-B/ViTAEv2 with RVSA; 100 M. & MillionAID optical imagery. \textbf{[O]} & 0.5--153 & SN & FT $\rightarrow$ Cls., Det., Seg. \textbf{I}: Multiscale landscape stratification; snapshot proxy evaluation provides no process test. \\
\mbox{Nov. 2022} & Satlas\newline Pretrain \cite{bastani_satlaspretrain_2023} & SatlasNet/Swin; 88 M. & NAIP and Sentinel-2 imagery with multisource label products. \textbf{[M]} & 1--10 & SN & FT $\rightarrow$ Cls., Det., Seg., CD. \textbf{I}: Supplies land-surface context, but no direct ecohydrological state or flux was evaluated. \\
\mbox{Dec. 2022} & Scale-MAE \cite{reed_scale-mae_2023} & ViT-L; 322.9 M. & fMoW RGB imagery and resolution metadata. \textbf{[O]} & n/r & SN & LP/FT $\rightarrow$ Cls., Seg. \textbf{I}: Scale-aware transfer may reduce mapping mismatch; temporal and process-sensitive measurements are absent. \\
\addlinespace[2pt]
\mbox{Feb. 2023} & GFM \cite{mendieta_towards_2023} & Swin-B; 88 M. & GeoPile high- and medium-resolution optical datasets. \textbf{[O]} & 0.1--30 & SN & FT $\rightarrow$ Cls., Seg., CD, SR. \textbf{I}: Broad spatial transfer, with semantic rather than process-based validation. \\
\mbox{Apr. 2023} & BFM \cite{cha_billion-scale_2026} & Parallel-block ViT; up to 2.42 B. & MillionAID optical imagery. \textbf{[O]} & 0.5--153 & SN & FT $\rightarrow$ Det., Seg. \textbf{I}: Large optical capacity improves mapping, but snapshots and proxy labels do not establish physical skill. \\
\mbox{Apr. 2023} & Presto \cite{tseng_lightweight_2024} & Lightweight transformer; 0.402 M. & Sentinel-1/2, ERA5, DEM and derived monthly variables. \textbf{[B]} & 10 & TS & FE/FT $\rightarrow$ Cls., Reg. \textbf{D-I}: LFMC regression against 1,578 field samples under a geographically separated test; the task-specific random forest remained stronger. \\
\mbox{Jun. 2023} & SSL4EO-L \cite{stewart_ssl4eo-l_2023} & ResNet/ViT; up to 86 M. & Five million multiseason Landsat 4--9 patches. \textbf{[O]} & 30 & PS & FT $\rightarrow$ Seg., cloud mapping. \textbf{I}: Supports long-record applications, but only short sequences and indirect tasks were tested. \\
\mbox{Sep. 2023} & EarthPT \cite{smith_earthpt_2024} & Decoder-only transformer; 700 M. & ClearSky-derived Sentinel-2-equivalent reflectance sequences. \textbf{[O]} & 10 & TS & FE/FT $\rightarrow$ reflectance/NDVI Fcst., land-use Cls. \textbf{I}: Enables trajectory analysis; direct water--carbon inference was not tested. \\
\mbox{Oct. 2023} & Prithvi-EO-1.0 \cite{jakubik_foundation_2023} & 3-D ViT; 100 M. & Harmonized Landsat--Sentinel-2 reflectance sequences. \textbf{[O]} & 30 & PS & LP/FT $\rightarrow$ Cls., Seg. \textbf{I}: Supports flood, crop and land-cover mapping, but not direct states or fluxes. \\
\mbox{Nov. 2023} & CROMA \cite{fuller_croma_2023} & Dual/joint ViT; 86 M. & One million matched Sentinel-1 GRD and Sentinel-2 L2A samples. \textbf{[M]} & 10 & PS & LP/FT $\rightarrow$ Cls., Seg. \textbf{I}: Cloud-robust wetness and inundation context; no independent continuous-state validation. \\
\mbox{Nov. 2023} & Spectral\newline GPT \cite{hong_spectralgpt_2024} & 3-D spectral transformer; up to 600 M. & Sentinel-2 multispectral imagery. \textbf{[O]} & 10--60 & SN & LP/FT $\rightarrow$ Cls., Seg., CD. \textbf{I}: Vegetation and composition information, without temporal or direct process validation. \\
\mbox{Dec. 2023} & FoMo \cite{bountos_fomo_2025} & ViT with band-wise tokens; 110 M. & Satellite, aerial and forest-inventory sources spanning optical, SAR and DEM. \textbf{[B]} & 10--20 & PS & LP/FT $\rightarrow$ Cls., Seg., Det. \textbf{I}: Forest structure/disturbance context; water and carbon states or fluxes were not independently tested. \\
\mbox{Dec. 2023} & SkySense \cite{guo_skysense_2024} & Factorised multimodal spatiotemporal transformer; 2.06 B. & High-resolution optical and Sentinel-1/2 time series. \textbf{[M]} & 0.3--10 & TS & FE/FT $\rightarrow$ Cls., Det., Seg., CD. \textbf{I}: Crop, flood and land-cover evidence remains based on semantic labels. \\
\addlinespace[2pt]
\mbox{Mar. 2024} & DOFA \cite{xiong_neural_2025} & Wavelength-conditioned dynamic ViT; up to 337 M. & Sentinel-1/2, Gaofen, NAIP, EnMAP and related sensors. \textbf{[M]} & 1--30 & SN & FT/PEFT $\rightarrow$ Cls., Seg., Det., Reg. \textbf{I}: Supports vegetation, soil and water mapping; no temporal or direct process constraint. \\
\mbox{Mar. 2024} & MTP \cite{wang_mtp_2024} & Shared CNN/ViT encoder with task decoders; $>$300 M. & SAMRS optical imagery and segmentation/detection labels. \textbf{[O]} & 0.3--30 & SN & FT $\rightarrow$ Cls., Det., Seg., CD. \textbf{I}: Broad semantic transfer, but snapshots do not resolve ecohydrological dynamics. \\
\mbox{Mar. 2024} & SatMAE++ \cite{noman_rethinking_2024} & ViT-L with convolutional upsamplers; 307 M. & fMoW and fMoW-Sentinel optical/multispectral imagery. \textbf{[O]} & 10--60 & PS & LP/FT $\rightarrow$ Cls. \textbf{I}: Multiscale spectral context; no process-scale memory or direct physical validation. \\
\mbox{Apr. 2024} & msGFM \cite{han_bridging_2024} & Swin transformer; 89 M. & Two million paired/unpaired images from four optical and microwave sensors. \textbf{[M]} & 0.1--153 & SN & FT $\rightarrow$ Cls., Seg., cloud removal, pan-sharpening. \textbf{I}: Cross-sensor reuse is demonstrated; state and flux skill is not. \\
\mbox{Apr. 2024} & OmniSat \cite{astruc_omnisat_2025} & Heterogeneous encoders and fusion transformer; n/r. & Aligned aerial RGB, Sentinel-1/2 time series and ancillary context. \textbf{[B]} & 10 & TS & LP/FT $\rightarrow$ forestry, land-cover and crop Cls./Seg. \textbf{I}: No direct water, energy or carbon evaluation. \\
\mbox{May 2024} & MMEarth \cite{nedungadi_mmearth_2025} & ConvNeXt-V2; 3.7 M. & Sentinel-2 paired with Sentinel-1, DEM, climate and land-cover variables. \textbf{[B]} & 10 & TS & LP/FT $\rightarrow$ Cls., Seg. \textbf{I}: Environmental context is encoded, but derived inputs can make product-target evaluation circular. \\
\mbox{May 2024} & SatSwin\newline MAE \cite{nakayama_satswinmae_2024} & Hierarchical Video Swin; n/r. & Multiscale Sentinel-2 image time series. \textbf{[O]} & 10--60 & TS & FT $\rightarrow$ Seg., Reg., flood/fire/crop mapping. \textbf{I}: Phenology/change utility is shown; physical targets were not evaluated. \\
\mbox{Aug. 2024} & Spectral\newline Earth \cite{braham_spectralearth_2025} & Spectral-adapted ResNet/ViT; n/r. & Globally distributed EnMAP hyperspectral patches. \textbf{[O]} & 30 & SN & LP/PEFT/FT $\rightarrow$ Cls., Seg., Reg. \textbf{I}: May constrain plant and soil traits, but remains snapshot-based and indirectly evaluated. \\
\mbox{Oct. 2024} & CrossEarth \cite{gong_crossearth_2026} & ViT with Earth-style injection; 86 M. & Diverse optical segmentation domains, sensors and styles. \textbf{[O]} & 0.5--153 & SN & FT $\rightarrow$ domain-generalised Seg. \textbf{I}: Supports geographic transfer; continuous states and fluxes were not tested. \\
\mbox{Nov. 2024} & Clay v1.5 \cite{clay_foundation_clay_2024} & ViT-L/16; 311 M encoder. & Sentinel-2, Landsat, NAIP, LINZ, MODIS and Sentinel-1 plus metadata. \textbf{[B]} & 0.5--500 & PS & FE/FT $\rightarrow$ Cls., Seg., Reg. \textbf{I}: Biomass/change utility; short sequences and weak extreme-event coverage limit process inference. \\
\mbox{Dec. 2024} & AnySat \cite{astruc_anysat_2025} & Scale-adaptive multimodal encoder; 127 M. & Eleven optical and SAR sensors spanning varied scales and cadences. \textbf{[M]} & 0.2--250 & TS & LP/FT $\rightarrow$ Cls., Seg., CD. \textbf{I}: Broad transfer, but thermal, passive-microwave and SIF constraints are absent. \\
\mbox{Dec. 2024} & Prithvi-EO-2.0 \cite{szwarcman_prithvi-eo-20_2026} & 3-D ViT; 300/600 M. & Global HLS sequences plus time and location metadata. \textbf{[B]} & 30 & PS & LP/FT $\rightarrow$ Cls., Seg., Reg. \textbf{D-I}: GPP partitioned from eddy-covariance NEE at 37 sites under leave-one-year-out validation; attribution remains shared with MERRA-2 forcing and NEE partitioning. \\
\mbox{Dec. 2024} & SeaMo \cite{li_seamo_2026} & Multimodal transformer; 86 M. & Multiseason Sentinel-1/2 and related optical--SAR datasets. \textbf{[M]} & 10 & TS & FT $\rightarrow$ Cls., Seg., CD. \textbf{I}: Phenology and change context; no direct state or flux evaluation. \\
\addlinespace[2pt]
\mbox{Feb. 2025} & Galileo \cite{tseng_galileo_2025} & Multimodal transformer; 85 M. & Sentinel-1/2, NDVI, DEM, ERA5 and TerraClimate/product variables. \textbf{[B]} & 10 & TS & LP/FT $\rightarrow$ Cls., Seg., Reg. \textbf{I}: Crop, flood and drought context; target-like inputs can make evaluation circular. \\
\mbox{Mar. 2025} & Copernicus-FM \cite{wang_towards_2025} & Dynamic-hypernetwork ViT; n/r. & Aligned Sentinel-1/2/3/5P and DEM samples. \textbf{[B]} & 10--1,000 & PS & FT $\rightarrow$ preprocessing, Cls., Seg., Reg. \textbf{I}: Broad land/atmosphere coverage; direct outputs are mainly product-derived. \\
\mbox{Mar. 2025} & FlexiMo \cite{li_fleximo_2026} & DOFA ViT-B with resolution/channel adapters; n/r. & Inherited DOFA multisensor inputs plus resolution/wavelength metadata. \textbf{[M]} & n/r & SN & FT $\rightarrow$ Cls., Seg., cloud detection. \textbf{I}: Flexible sensor transfer; spatial support and direct process validation remain unresolved. \\
\mbox{Mar. 2025} & Panopticon \cite{waldmann_panopticon_2025} & ViT-B with channel cross-attention; 98.1 M. & fMoW, SatlasPretrain, MMEarth and SpectralEarth sensors. \textbf{[B]} & 0.3--100 & SN & LP/FT $\rightarrow$ Cls., Seg. \textbf{I}: Broad sensor coverage, but snapshot evaluation does not establish process skill. \\
\mbox{Mar. 2025} & RoMA \cite{wang_roma_2025} & Vision Mamba; 297 M. & OpticalRS-4M and MillionAID optical images. \textbf{[O]} & n/r & SN & LP/FT $\rightarrow$ Cls., Seg., CD. \textbf{I}: Scalable spatial representation; spatial support and direct ecohydrological tests are absent. \\
\mbox{Apr. 2025} & Complex\newline SAR-FM \cite{wang_complex-valued_2026} & Complex-valued Swin; n/r. & Complex and general SAR imagery. \textbf{[M]} & n/r & SN & FT $\rightarrow$ Cls., Det., Seg., CD. \textbf{I}: Scattering-aware surface mapping; optical, temporal and direct process constraints are absent. \\
\mbox{Apr. 2025} & PyViT-FUSE \cite{weber_pyvit-fuse_2025} & Pyramidal ViT with attention fusion; 103 M. & Co-registered SPOT, Sentinel-1/2 and Landsat-8 bands. \textbf{[M]} & 1.5--30 & PS & FE/FT $\rightarrow$ Seg. and mixed-band transfer. \textbf{I}: Robust mapping, but short temporal support and indirect labels limit process inference. \\
\mbox{Apr. 2025} & RingMoE \cite{bi_ringmoe_2026} & Hierarchical mixture of experts; 14.7 B, prunable to 1 B. & 400 million optical, multispectral and SAR images from nine satellites. \textbf{[M]} & n/r & SN & FT $\rightarrow$ Cls., Det., Seg., tracking, CD, depth. \textbf{I}: Broad semantic capacity; spatial support and physical validation are absent. \\
\mbox{Apr. 2025} & TerraMind \cite{jakubik_terramind_2025} & ViT encoder--decoder; n/r. & TerraMesh with Sentinel-1/2, DEM, NDVI, land cover, captions and location. \textbf{[B]} & 10 & PS & ZS/FT $\rightarrow$ Cls., Seg., Reg., Gen. \textbf{I}: Missing-mode generation is useful, but does not ensure radiometric or process consistency. \\
\mbox{May 2025} & AgriFM \cite{li_agrifm_2026} & Video Swin; n/r. & MODIS, Landsat-8/9 and Sentinel-2 surface reflectance. \textbf{[O]} & 10--500 & TS & FT $\rightarrow$ crop/field Cls. and Seg. \textbf{I}: Stratifies phenology/management; no water, energy or carbon state or flux was validated. \\
\mbox{May 2025} & TiMo \cite{qin_timo_2025} & Hierarchical ViT with spatiotemporal gyroscope attention; up to 675 M. & MillionST Sentinel-2 sequences over five years. \textbf{[O]} & 10 & TS & FT $\rightarrow$ Seg., CD. \textbf{I}: Phenology/disturbance context; sparse sampling limits event-scale sensitivity. \\
\mbox{Jun. 2025} & CGEarth\newline Eye \cite{yi_cgeartheyehigh-resolution_2025} & ViT-S/B/L/H/G; up to 1.1 B. & Jilin-1 quarterly sub-metre optical imagery. \textbf{[O]} & 0.75 & PS & FE/FT $\rightarrow$ Cls., Det., Seg., CD. \textbf{I}: Local monitoring at high spatial detail; semantic tasks do not validate continuous variables. \\
\mbox{Jun. 2025} & TerraFM \cite{danish_terrafm_2025} & Multisensor ViT with cross-attention; 300 M. & Globally distributed Sentinel-1/2 imagery. \textbf{[M]} & 10 & TS & FE/FT $\rightarrow$ Cls., Seg. \textbf{I}: Dynamic land-surface context; direct state/flux validation is absent. \\
\mbox{Jun. 2025} & TESSERA \cite{feng_tessera_2026} & Pixel-wise temporal encoder; n/r. & Sentinel-2 surface-spectra time series. \textbf{[O]} & 10 & AP & FE/light heads $\rightarrow$ Cls., Seg., Reg. \textbf{I}: Annual embeddings support monitoring but cannot establish event sensitivity or process memory. \\
\mbox{Jul. 2025} & AlphaEarth\newline Foundations \cite{brown_alphaearth_2025} & Teacher--student video embedding-field model; ${\sim}$480 M. & Optical, SAR, lidar and geospatial/text context. \textbf{[B]} & 10 & AP & FE/LP $\rightarrow$ Cls., Reg. \textbf{D-P}: OpenET ET regression uses a retrieved product; annual embeddings do not establish independent flux skill. \\
\mbox{Jul. 2025} & SkySense V2 \cite{zhang_skysense_2025} & Unified multimodal transformer/MoE; 665 M. & High-resolution optical and Sentinel-1/2 time series. \textbf{[M]} & 0.3--10 & TS & FT $\rightarrow$ Cls., Det., Seg., CD. \textbf{I}: Broad land monitoring, with evidence remaining indirect. \\
\mbox{Aug. 2025} & SkySense\newline ++ \cite{wu_semantic-enhanced_2025} & Factorised multimodal spatiotemporal encoder; 2.06 B. & 27 million high-/medium-resolution optical and SAR images. \textbf{[M]} & 0.05--10 & TS & Few-shot/FT $\rightarrow$ Cls., Det., Seg. \textbf{I}: Crop, tree and flood labels do not validate continuous states or fluxes. \\
\mbox{Sep. 2025} & FUSAR-KLIP \cite{yang_fusar-klip_2025} & SAR--language/image dual encoder; n/r. & SAR imagery paired with text and task-derived semantic descriptions. \textbf{[M]} & n/r & PS & ZS/FT $\rightarrow$ Cls., Det., Seg., Ret., caption/VQA. \textbf{I}: Semantic retrieval lacks optical, temporal and process constraints. \\
\mbox{Sep. 2025} & StefaLand \cite{kraabel_stefaland_2026} & Attribute transformer with residual adapters; n/r. & Static landscape attributes and meteorological/land-surface time series. \textbf{[B]} & n/r & TS & PEFT/FT $\rightarrow$ streamflow, soil-moisture/composition Reg., landslide Cls. \textbf{D-I}: \emph{In situ} SM and observed streamflow evaluated under spatial holdouts; architecture, broader inputs and pretraining contributions remain difficult to separate. \\
\mbox{Nov. 2025} & OlmoEarth \cite{herzog_olmoearth_2026} & ViT family; up to 300 M. & Sentinel-1/2, Landsat-8, DEM, land, crop, canopy and OSM layers. \textbf{[B]} & 10 & TS & FE/LP/FT $\rightarrow$ Cls., Seg. \textbf{I}: Crop, flood and canopy utility; derived-map inputs may encode target information. \\
\mbox{Dec. 2025} & Any-Optical-Model \cite{li_any-optical-model_2025} & Any-band multiscale ViT; n/r. & Sentinel-2, Landsat, HLS and other optical sensors. \textbf{[O]} & 0.1--100 & SN & FT $\rightarrow$ Cls., Seg., CD, object extraction. \textbf{I}: Vegetation/surface mapping without microwave, thermal or direct process constraints. \\
\addlinespace[2pt]
\mbox{Jan. 2026} & THOR \cite{forgaard_thor_2026} & Compute-adaptive ViT; 314.4 M. & Sentinel-1/2/3 OLCI--SLSTR, ERA5 and map products. \textbf{[B]} & 10--1,000 & TS & FE/FT $\rightarrow$ Cls., Seg., Reg. \textbf{I}: Broad environmental context; mixed scales and indirect benchmarks do not establish process validity. \\
\mbox{Mar. 2026} & CrossEarth-SAR \cite{ye_crossearth-sar_2026} & Physics-guided sparse MoE; 1 B. & CrossEarth-SAR-200K imagery and physical SAR descriptors. \textbf{[M]} & n/r & SN & FT $\rightarrow$ domain-generalised Seg. \textbf{I}: Supports inundation/surface mapping; no direct state or flux evaluation. \\
\mbox{Mar. 2026} & SIGMAE \cite{zhang_sigmae_2026} & ViT encoder--decoder; n/r. & Multispectral imagery and derived spectral-index priors. \textbf{[O]} & 10--60 & SN & FT $\rightarrow$ Cls., Seg., object extraction, CD. \textbf{I}: Vegetation/water mapping remains based on snapshot proxy tasks. \\
\mbox{Mar. 2026} & TerraFlow \cite{puriy_terraflow_2026} & TerraMind encoder--decoder with temporal attention; n/r. & Variable-length aligned optical--SAR sequences. \textbf{[M]} & 10 & TS & FT $\rightarrow$ temporal Cls./Seg./CD and flood/wildfire-risk Reg. \textbf{I}: Captures land-surface dynamics; continuous water--carbon variables lack independent validation. \\
\mbox{Apr. 2026} & HighFM \cite{girtsou_highfm_2026} & ViT-B; ${\sim}$90 M. & More than 2 TB of 11-band SEVIRI radiances. \textbf{[B]} & 3,000 & PS & FT $\rightarrow$ cloud Seg., active-fire Det. \textbf{I}: Sub-hourly thermal context is relevant to LST/ET/heat stress; demonstrations focus on clouds/fire. \\
\mbox{May 2026} & FLORO \cite{rodriguez_floro_2026} & Availability-aware ViT; n/r. & Sentinel-1/2, SkySat, DEM and UAV-derived products. \textbf{[B]} & n/r & TS & FE $\rightarrow$ Cls., Seg., Reg. \textbf{I}: Ecological transfer is relevant, but independent state/flux validation remains unverified. \\
\mbox{May 2026} & HySens \cite{zhao_hysens_2026} & 1-D spectral transformer with wavelength harmoniser; n/r. & 188 million EnMAP hyperspectral reflectance spectra. \textbf{[O]} & 30 & SN & FE/FT $\rightarrow$ Cls., soil/water-property Reg. \textbf{D-I}: Field-referenced bare-soil moisture, soil organic carbon and water chlorophyll; direct but narrow and snapshot-based. \\
\mbox{May 2026} & Spectral\newline Earth-FM \cite{braham_spectralearth-fm_2026} & Hierarchical sensor-specific transformer with cross-sensor fusion; n/r. & EnMAP, EMIT, DESIS, Sentinel-2, Landsat optical/LST and Sentinel-1. \textbf{[B]} & 10--100 & PS & LP/FT $\rightarrow$ Cls., Seg., Reg. \textbf{I}: Traits, LST and moisture-sensitive observations; no SIF pathway or independent process validation. \\
\mbox{Jul. 2026} & Mini-JEPAs \cite{rahman_sensor-specialized_2026} & Five ViT-S specialists; 22 M each. & Sentinel-2 optical/phenology, Sentinel-1 SAR, MODIS thermal and topography--soil stack. \textbf{[B]} & 30 & TS & FE/LP plus routed Ret./Reg. \textbf{D-P}: Improves SMAP soil moisture, aridity and precipitation-product prediction; hydrological decisions remain untested. \\
\mbox{Jul. 2026} & TESSERA v2 \cite{feng_tessera_2026-1} & Pixel-wise encoder; 1 B teacher/21 M student. & Irregular Sentinel-1/2 time series. \textbf{[M]} & 10 & AP & FE/light heads $\rightarrow$ Cls., Seg., Reg., CD. \textbf{I}: Scaled annual embeddings improve mapping, but retain annual compression and lack direct process validation. \\

\end{longtable}

\normalsize
\begin{multicols}{2}
\subsection{Synthesis of Model-Level Capabilities}
\label{sec:meta_synthesis}

Taken together, the 60 releases describe a heterogeneous EOFM design space across backbone architecture, parameter scale, observation pathways, spatial and temporal support, adaptation and downstream outputs. Backbones range from lightweight convolutional and transformer encoders to multimodal fusion, encoder--decoder and mixture-of-experts architectures. Reported parameter counts span 0.402 million to 14.7 billion, with a median of 300 million across 43 releases, while representative pre-training sampling distances extending from approximately 1.7~m to 3~km. Temporal designs include snapshots, paired observations or short sequences, explicit time series and annual embedding products. Fine-tuning is the predominant adaptation route, and release-paper evaluations concentrate on classification and segmentation. The observation pathways remain dominated by reflected optical and active-microwave data; thermal pathways are uncommon, and explicit passive-microwave-emission and SIF pathways are absent. Collectively, these characteristics show that current EOFMs offer flexible model scales, spatiotemporal supports and adaptation interfaces, with uneven coverage of ecohydrologically important observations and outputs.

Within the meta-analysis corpus, AlphaEarth Foundations, HySens and Mini-JEPAs provide the clearest direct relevance to ecohydrology because each evaluates an ecohydrological variable. AlphaEarth Foundations provides analysis-ready annual multisource embeddings and evaluates sparse-label transfer to monthly OpenET ET, supporting product-referenced spatial prediction at 10~m \cite{brown_alphaearth_2025}. HySens provides the strongest reference directness through independent field measurements of surface SM, SOC and water chlorophyll, although its evaluation remains narrow and snapshot-based \cite{zhao_hysens_2026}. Mini-JEPAs combine sensor-specialised optical, SAR, thermal and static environmental representations in an explicit time-series design and evaluate predictions of SMAP SM, aridity and precipitation products \cite{rahman_sensor-specialized_2026}. These releases provide complementary strengths in field-referenced property retrieval, analysis-ready spatial context and multisensor water--climate prediction. Their heterogeneous validation designs support target-specific selection; a general model ranking requires common targets, reference data, splits and metrics.

\end{multicols}

\twocolumn
\normalsize

A more specific application-level assessment is required to determine how these model capabilities translate to ecohydrological inference. Such an assessment connects each EOFM to target observability, reference independence, representation use, additional EO and meteorological inputs, validation or distribution-shift design, baseline comparisons, reported performance and principal limitations. Accordingly, we extend the review from model releases to application-centric research and trace the evidence from comparatively observable surface quantities through latent states and fluxes to events and decision-relevant outcomes. This application-centric perspective establishes which EOFM capabilities have been demonstrated and the evidentiary conditions governing those demonstrations.

\section{Capabilities of EOFMs across Ecohydrological Tasks}
\label{sec:eofm_capabilities}

The preceding meta-analysis established the breadth of the EOFM design space and identified three releases with direct ecohydrological evaluations, while their heterogeneous targets and validation designs preclude a general ranking of model capability. We now move from model-level evidence to application-level evidence, asking what individual studies demonstrate under their stated reference data, adaptation regimes, baseline comparisons and distribution shifts. Following the ecohydrological inference hierarchy in Fig.~\ref{fig:01_ecohydrol_hierarchy}, the synthesis is organised by inference level: T1 covers retrievals and indices, T2 encompasses estimated states and fluxes, and T3 addresses mechanisms, events and decision-relevant outcomes. We also distinguish model-release demonstrations, independent applications, comparative benchmarks and provisional grey literature because these study roles carry different risks of selective reporting and model-developer dependence. This structure establishes how far each application advances along the observation-to-inference pathway and the evidentiary conditions governing that advancement.

\begin{figure*}[!b]
\centering
\includegraphics[width=1.\textwidth]{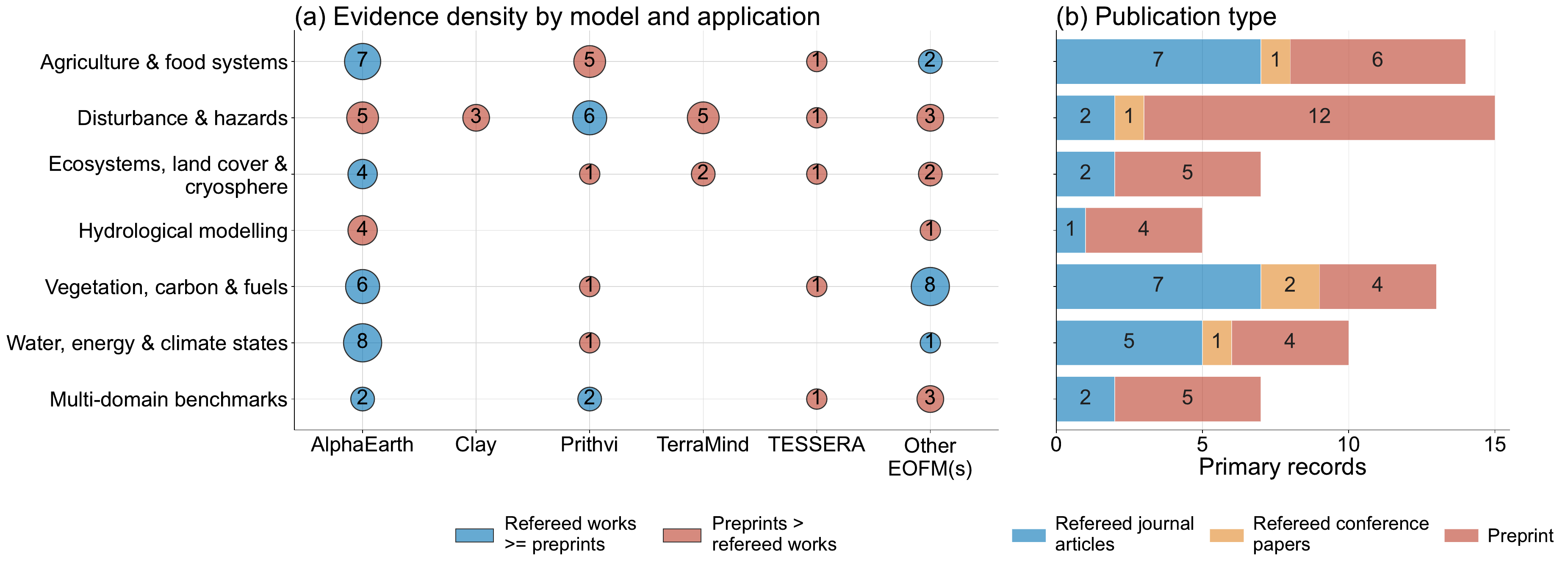}
\caption{Structure of the 71 eligible primary EOFM application records. (a) Evidence density by model and application family. Circle area and labels show record counts, while colour indicates whether refereed works are at least as numerous as preprints within each model--application combination. (b) Numbers of refereed journal articles, refereed conference papers and preprints by application family. A work with multi-model records contribute to each applicable model group in (a); therefore, the sum of record counts within each row of (a) may exceed the corresponding application-family count in (b).}
\label{fig:eofm_application_evidence}
\end{figure*}

\subsection{Overview of EOFM Applications}
\label{sec:eofm_application_overview}

To evaluate this application-level evidence, we conducted a targeted critical review using exact-model searches paired with LST, SM, vegetation, live fuel moisture content (LFMC), groundwater, ET, GPP, biomass, SOC, streamflow, flood, wildfire, ecological-trait and crop yield terms. We verified eligibility against the acquired full texts and publisher, proceedings, DOI or preprint metadata. The final audit contained 71 eligible primary-evidence records.

For each record, we extracted the model, application family, publication type, representation use, reference support, validation or distribution-shift design, baseline comparisons, result direction and principal limitations. Fig.~\ref{fig:eofm_application_evidence} summarises the model, application and publication structure, which shows broad but uneven application coverage. The corpus comprises agriculture and food systems (14 records), disturbance and hazards (15), ecosystems, land cover and cryosphere (7), hydrological modelling (5), vegetation, carbon and fuels (13), water, energy and climate states (10), and multi-domain benchmarks (7). AlphaEarth appears across all 7 application families. Publication maturity also varies: refereed works outnumber preprints in agriculture and food systems, vegetation, carbon and fuels, and water, energy and climate states, whereas preprints predominate in the other four families. These counts describe the composition of the curated corpus and the maturity of its evidence clusters.

The tiered synthesis progresses from retrievals and indices (T1), through estimated states and fluxes (T2), to mechanisms, events and decision-relevant outcomes (T3). Across these tiers, we assess the contribution of EOFM representations relative to accompanying observations, covariates and baselines, and the extent to which reference data, validation design and distribution-shift tests support the claimed ecohydrological inference.

\subsection{T1: Retrievals and Indices}
\label{sec:eofm_application_t1}

T1 applications evaluate whether EOFM representations support retrievals and indices that remain comparatively close to EO observables. The available evidence is concentrated in LST and SM, with studies spanning recovery from dynamic satellite observations, transfer to field spectroscopy and probes against gridded reference products. These cases allow the contribution of EOFM representations to be assessed alongside contemporaneous observations, handcrafted predictors and task-specific models.

\subsubsection{LST}

Dynamic thermal observations remain central to LST recovery, while EOFM representations can supply complementary spatial context. Lee-Burkhart \emph{et al.} fused AlphaEarth embeddings with GOES-18 spectral and auxiliary predictors to recover 5-min LST at 53 Hawai`i Mesonet stations \cite{lee-burkhart_geospatial_2026}. Leave-one-station-out RMSE decreased from 3.11 to 2.87~K, with the largest improvement under cloud.

\begin{figure*}[!b]
\centering
\includegraphics[width=1.\textwidth]{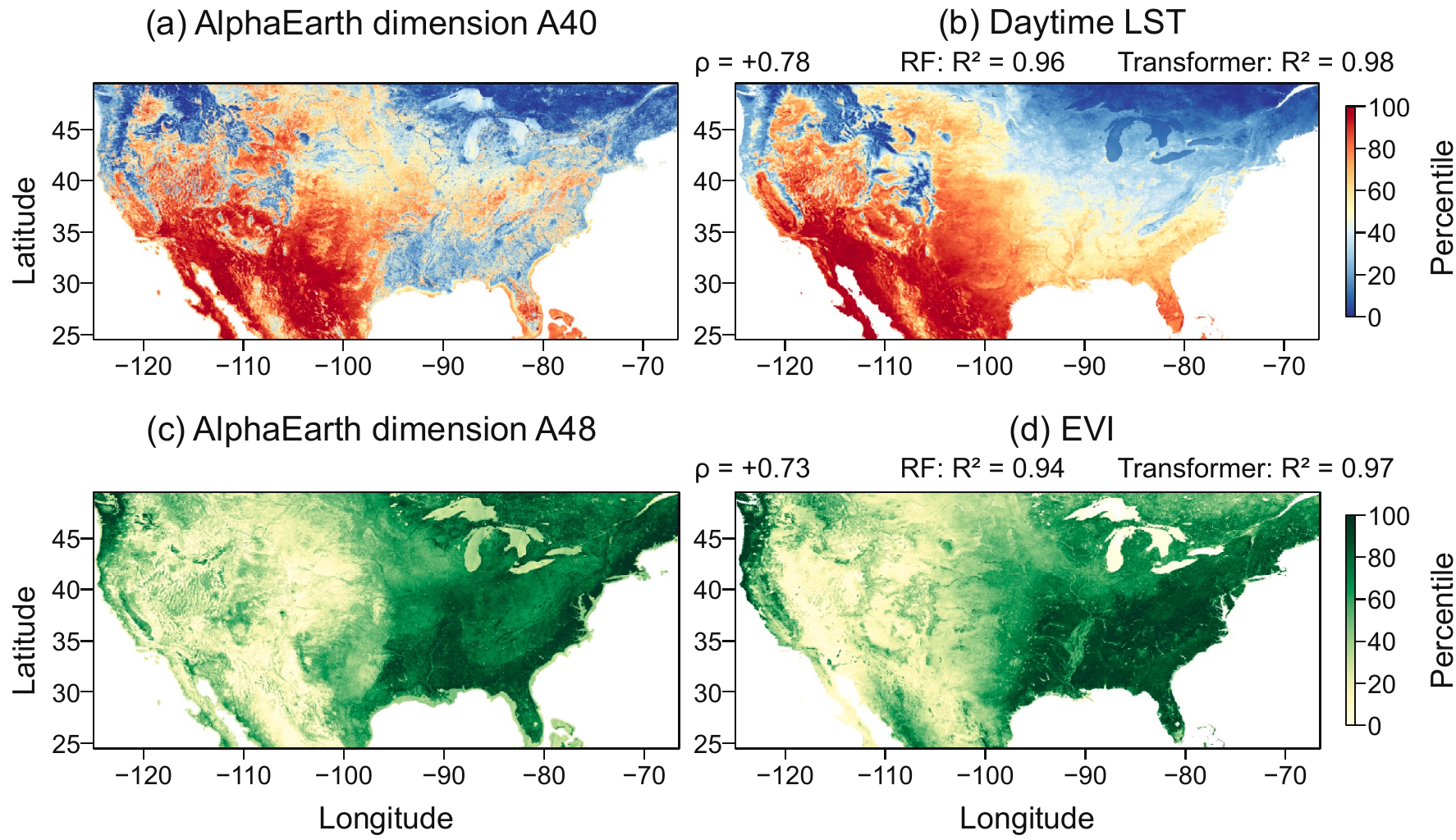}
\caption{Spatial correspondence between selected AlphaEarth embedding dimensions and environmental products across CONUS. Panels (a) and (b) compare AlphaEarth dimension A40 with daytime LST, and panels (c) and (d) compare dimension A48 with EVI. Mapped values are expressed as percentile ranks from 0 to 100. Spearman correlation ($\rho$), Random Forest $R^2$ and Transformer $R^2$ quantify the corresponding relationships. The paired maps demonstrate geographically coherent information within the representation and agreement with the selected products; independent recovery of the underlying environmental quantities requires separate reference observations. Reprinted from \cite{rahman_physically_2026} under a CC BY 4.0 License.}
\label{fig:eofm_application_t1_lst_evi}
\end{figure*}

Annual embeddings can encode terrain, land cover and climatological surface properties that support recovery when thermal observations are incomplete; the GOES-18 observations retain the sub-hourly temperature trajectory. Embedding probes provide complementary, product-referenced evidence. AlphaEarth probes reconstructed gridded thermal, vegetation and hydrological products with high accuracy and spatial stability across the contiguous United States (CONUS) (Fig.~\ref{fig:eofm_application_t1_lst_evi}) \cite{rahman_physically_2026}. These probes establish that the representation contains information associated with the selected products. Independent retrieval of LST requires separate reference observations because the gridded products carry their own retrieval and model histories \cite{yu_generating_2023}. Operational evaluation should therefore compare thermal harmonisation, conventional spatiotemporal gap filling, local ancillary variables and embeddings under station, heat island, season and extreme-heat holdouts.

\subsubsection{SM}

SM evidence spans field-spectroscopy retrieval, multisensor prediction and spatial-transfer experiments. HySens transferred wavelength-aware representations to field spectroscopy for bare-soil SM (and to co-located SOC and water-chlorophyll targets), with explicit scratch, frozen and fine-tuned comparisons \cite{zhao_hysens_2026}. This study provides rare direct evidence that pretraining can support small spectroscopy datasets. Its evaluation remains limited to snapshot measurements and does not test spatial neighbourhoods, temporal memory or satellite-to-field scaling.

The broader SM evidence includes positive, null and product-dependent results. A pan-European experiment combining Sentinel-1, Sentinel-2 and ERA5 found almost no incremental benefit from Prithvi features over handcrafted predictors under spatial cross-validation ($R^2=0.515$ versus 0.514) \cite{kontogiorgakis_comparative_2026}. The task-specific SAR--optical--weather GFM of Hashemi \emph{et al.} achieved high field-referenced skill for crop vegetation water content (VWC) and height across corn and soybean campaigns \cite{hashemi_estimating_2025}. Sensor-specialised Mini-JEPAs added a small but statistically stable increment to AlphaEarth for SMAP SM ($\Delta R^2=0.031$), although both feature sets degraded in leave-one-region-out SM tests \cite{rahman_sensor-specialized_2026}. StefaLand provides the strongest dynamic example because it uses meteorological and land-surface time series and evaluates global in situ SM under spatial holdouts; its architecture, broader inputs and pretraining contribution remain difficult to separate \cite{kraabel_stefaland_2026}. Together, these studies show that SM skill depends on access to relevant microwave observations, meteorological forcing and temporal information, as well as the persistence of that information under spatial transfer. The available evidence does not support a general claim that foundation features improve SM retrieval.

\subsection{T2: Estimated States and Fluxes}
\label{sec:eofm_application_t2}

T2 applications extend EOFM evaluation to latent land-surface states, ecosystem properties and fluxes that require stronger reference and modelling assumptions than T1 retrievals. The available evidence covers biomass, canopy biochemistry, ET, GPP, groundwater, LFMC, SOC and related soil properties, streamflow and water-quality indicators. Across these variables, EOFM representations most consistently provide transferable spatial context, while temporal forcing, local observations, adaptation strategy and reference directness govern performance and geographic transfer.

\subsubsection{Biomass}

\begin{figure*}[!b]
\centering
\includegraphics[width=1.\textwidth]{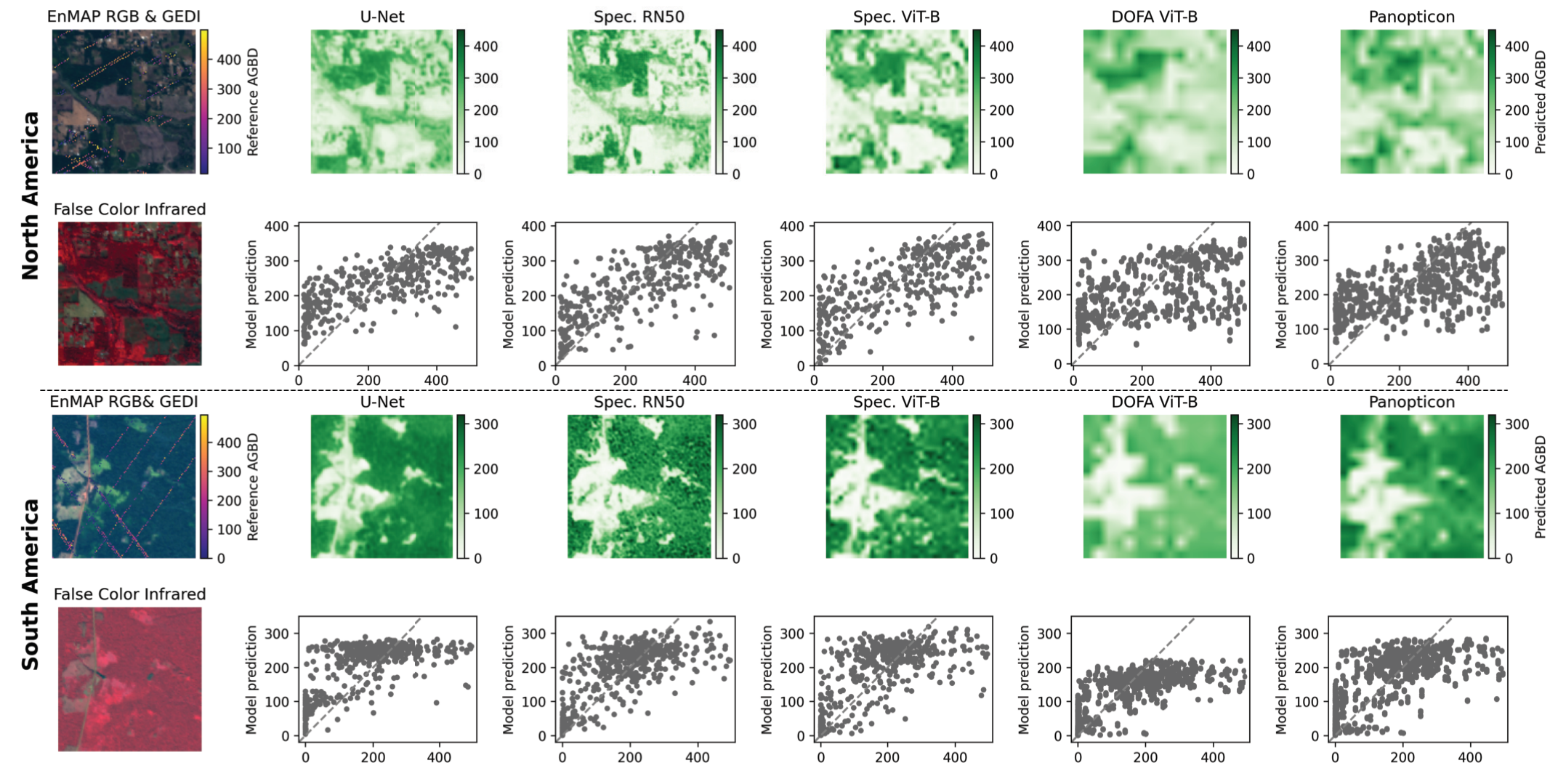}
\caption{Qualitative comparison of forest above-ground biomass density (AGBD) estimates for representative EnMAP hyperspectral image patches from North America (top two rows) and South America (bottom two rows). Within each regional block, the first column shows an EnMAP RGB composite overlaid with GEDI-derived reference AGBD values at lidar-shot locations and an infrared false-colour composite. The remaining columns present spatial predictions and reference--prediction scatterplots for the supervised U-Net baseline and four geospatial FMs evaluated with fine-tuned encoders: Spectral-ResNet-50 (Spec. RN50), Spectral-ViT-B, DOFA ViT-B and Panopticon. Dashed lines indicate 1:1 agreement, and AGBD is expressed in Mg~ha$^{-1}$. The GEDI-derived reference values are retrieved biomass-density estimates; independent field measurements are absent from this qualitative comparison. Reprinted from \cite{banze_hybiomass_2025} under a CC BY 4.0 License.}
\label{fig:eofm_application_t2_biomass}
\end{figure*}

Biomass provides the largest evidence base and the greatest variation in reference type, sensor support and adaptation strategy. Radar--optical contrastive features were linearly related to GEDI biomass, although satellite and biomass acquisitions were not necessarily contemporaneous \cite{prexl_multi-modal_2023}. TESSERA annual embeddings with lightweight heads matched strong task-specific baselines for LiDAR-referenced canopy height, Finnish BioMassters AGB and field-based agroforestry stocking indices, with high label efficiency \cite{feng_applications_2026}. Coupling the same embeddings with attentive neural processes retained competitive accuracy, improved the calibration of cross-biome GEDI prediction intervals and enabled adaptation with a small local sample \cite{young_interpolation_2026}. A global comparison found substantially weaker performance for runnable frozen encoders than for a supervised baseline, while AlphaEarth features produced the strongest results and transferred more effectively across space and time \cite{sialelli_above-ground_2026}. HyBiomass and FLORO broaden sensor and scale coverage using retrieved biomass products or benchmark labels as references \cite{banze_hybiomass_2025,rodriguez_floro_2026}. Fig.~\ref{fig:eofm_application_t2_biomass} illustrates how a supervised U-Net and four fine-tuned geospatial FMs reproduce AGBD patterns from EnMAP imagery across North and South American examples. Their differing spatial smoothness and departures from the 1:1 line demonstrate model- and region-dependent prediction behaviour. The GEDI-derived reference provides a common retrieved target for this comparison; independent field validation is absent.

Independent AlphaEarth studies further delimit this evidence. Spectral indices consistently outperformed AlphaEarth embeddings for field-referenced biomass in a regenerating Andean forest \cite{rojas-lucero_spectral_2026}. In the northeastern United States, a LiDAR--inventory--embedding workflow reached $R^2=0.79$ and 0.82 after growth-adjusted temporal augmentation; the embedding-specific contribution and uncertainty in the adjusted reference require separate evaluation \cite{lamahewage_foundation-model_2026}. MMEarth-Bench also found that pretraining advantages were greatest with limited labels and could disappear under abundant multimodal supervision \cite{gordon_mmearth-bench_2026}. Overall performance therefore depends strongly on canopy-structure information, LiDAR support, local calibration, adaptation and uncertainty modelling. Biomass maps estimate structural carbon stocks; current GPP, respiration, net ecosystem exchange and durable net carbon storage require additional flux and process evidence.

\subsubsection{Canopy Chlorophyll, Nitrogen and Phosphorus}

Canopy biochemical retrieval provides a narrower T2 use case. A peer-reviewed study selected compact subsets of AlphaEarth channels to retrieve canopy chlorophyll, nitrogen and phosphorus across 24 approximately one-hectare maize plots (about 87~m $\times$ 105~m each) at the Kellogg Biological Station in Michigan, USA \cite{alam_dimensionality_2025}. The result demonstrates label efficiency and vegetation information within the representation. Robust nutrient monitoring across regions, seasons and canopy structures requires broader sampling and independent transfer evaluation.

\subsubsection{ET}

ET evidence spans dynamic hybrid estimation and transfer from an operational product. A temporal Transformer combined annual AlphaEarth embeddings, Sentinel-2 NDVI and daily meteorology to produce 10-m urban ET estimates, reaching temporally held-out FLUXNET $R^2=0.92$ and RMSE$=0.37$~mm~d$^{-1}$; performance over the Shenzhen urban-flux observations was lower ($R^2=0.56$) \cite{jiang_transformer-based_2026}. Meteorological forcing supplied daily variability, while the embeddings represented spatial heterogeneity. AlphaEarth also achieved $R^2=0.58\pm0.01$ for monthly OpenET labels and was the only tested representation above $R^2=0.2$ in that evaluation \cite{brown_alphaearth_2025}. This result demonstrates sparse-label transfer of an operational ET product. \textit{OpenET} is an ensemble of satellite and meteorological models, and its uncertainty and modelling assumptions remain part of the reference pathway \cite{melton_openet_2022}.

\subsubsection{GPP}

GPP has been evaluated against daily flux-tower estimates with greater reference directness. Prithvi-EO-2.0 was tested against GPP partitioned from eddy-covariance net ecosystem exchange at 37 sites. Under leave-one-year-out evaluation, a fused Prithvi--MERRA-2 model averaged $R^2=0.81$, compared with 0.75 for a same-input ResNet \cite{szwarcman_prithvi-eo-20_2026}. The result supports hybrid prediction from EO representations and meteorological forcing, subject to uncertainty from NEE partitioning, tower-footprint mismatch, cloud filtering and the contribution of MERRA-2. The preservation of seasonal drought responses, GPP--ET coupling and water-use efficiency under ecosystem and extreme-event shifts remains untested.

\subsubsection{Groundwater}

Groundwater applications highlight the influence of hydrogeological context. AlphaEarth embeddings combined with the Tabular Prior-Data Fitted Network (TabPFN) predicted groundwater depth from 87 wells in the Leizhou Peninsula with $R^2=0.793$ \cite{li_new_2026}. Leave-one-out validation within this single hydrogeological setting leaves regional transfer unresolved. In Danish peatlands, an embedding-only model was 3\% less accurate than an expert-covariate model, while basic topographic information recovered part of the deficit and produced more plausible responses under synthetic wet and drained conditions \cite{koch_modelling_2026}. These studies support embeddings as contextual predictors whose transfer depends on hydrogeology, topography and regional sampling.

\subsubsection{LFMC}

LFMC provides a clear test of adaptation strategy. The Presto release evaluation compared frozen and fine-tuned time-series representations against 1,578 geographically separated field samples. Fine-tuning improved on random initialisation, while the task-specific random forest remained stronger \cite{tseng_lightweight_2024}. Johnson \emph{et al.} subsequently fine-tuned the more broadly multimodal Galileo encoder to produce spatially complete 10-m LFMC maps, reporting more than 20\% lower RMSE than random initialisation and demonstrations over two recent California fire regions \cite{johnson_high-resolution_2025}. Presto therefore tests representation content against a strong simple baseline, and Galileo tests whether pretraining supports a wall-to-wall mapping workflow. Evidence for fire behaviour, ignition probability and responses to unprecedented vegetation stress remains dependent on event-specific calibration and evaluation.

\subsubsection{SOC, Soil N and pH}

Soil-property evidence similarly favours representations combined with environmental covariates or local reference data. In Danish peatlands, the embedding-only SOC model was 6\% less accurate than the expert-covariate model, and adding topography recovered part of the deficit \cite{koch_modelling_2026}. Across more than 1,800 cropland topsoil samples, AlphaEarth embeddings outperformed bare-soil and vegetation-index approaches, with performance comparable to a spatial kriging--Cubist hybrid \cite{castaldi_monitoring_2026}. MMEarth-Bench extended the comparison to SOC, soil N and pH under random and Africa holdouts. For each task, all African tiles formed the geographic test set, and the remaining global tiles supplied the training, validation and random-test sets, thereby testing transfer to a continent underrepresented in field-labelled soil data \cite{gordon_mmearth-bench_2026}. Multimodal pretraining provided the largest gains when labels were scarce; geographic generalisation remained poor, and a randomly initialised multimodal model was competitive when labels were abundant \cite{gordon_mmearth-bench_2026}. Soil-forming factors and the spatial support of reference samples therefore remain central to transfer.

\subsubsection{Streamflow}

Streamflow applications use EOFM representations primarily as catchment descriptors. AlphaEarth embeddings improved donor-basin selection and spatially out-of-sample simulation, with gains declining as environmentally dissimilar donors were added \cite{qu_utilizing_2026}. Across 531 CAMELS-US catchments, the strongest pseudo-ungauged result combined embeddings with auxiliary static descriptors in an adapted LSTM; the study attributed the gain primarily to improved input data \cite{heudorfer_better_2026}. In 455 Australian basins, basin-mean AlphaEarth embeddings increased median NSE from 0.711 to 0.735 for temporal generalisation and from 0.609 to 0.649 for spatial generalisation, with the largest error reduction in isolated basins \cite{ou_foundation-scale_2026}. Fusion across a regulated river network increased median reconstruction NSE from 0.301 to 0.533, although annual embeddings weakly represented reservoir operations and other fast human controls \cite{lin_fusing_2026}. StefaLand extends this approach by pretraining jointly on static attributes and time-series forcing, then evaluating CAMELS, Caravan and global SM with spatial holdouts \cite{kraabel_stefaland_2026}. Collectively, these studies support learned catchment descriptors alongside meteorological forcing and hydrological memory. Causal relationships among storage, runoff generation and groundwater connectivity require water-balance and regime diagnostics.

\subsubsection{Water-Quality Indicators}

Chemical water-quality targets provide a boundary test for surface representations. Gain-ranked AlphaEarth channels predicted eight indicators across the South Korean national river network with leave-one-year-out correlations of 0.62--0.89 \cite{kim_estimating_2025}. Coordinates contributed strongly, shoreline artefacts persisted and spatial transfer was not evaluated. For seasonal farmland-groundwater nitrate, measured hydrochemistry exceeded embedding-based predictors by approximately 10--20\% in $R^2$; virtual-sample augmentation improved the use of embeddings, while subsurface observations retained the strongest predictive information \cite{xu_hydrochemistry_2026}. These cases establish compact predictive information associated with water chemistry. Independent observations remain necessary to characterise the underlying chemical state.

\subsection{T3: Mechanisms, Events, and Decision-Relevant Outcomes}
\label{sec:eofm_application_t3}

T3 applications extend EOFM evaluation to events, ecological composition and outcomes that can inform management or risk decisions. The available evidence covers crop type, stress, tillage and yield; ecological traits and composition; flood inundation; glacial-lake extent; landslide susceptibility; and wildfire burned area and severity. Most studies evaluate retrospective mapping or classification under geographic, temporal, event or sensor holdouts. Prospective decision benefit, causal process attribution and post-event recovery remain sparsely evaluated.

\subsubsection{Crop Type, Stress, Tillage and Yield}

Agricultural variables sit at the boundary between monitoring and decision support. County-aggregated AlphaEarth embeddings predicted corn and soybean yield under leave-one-year-out testing with mean $R^2$ values of 0.825 and 0.814, respectively \cite{fang_application_2026}. Strong spatial autocorrelation in the residuals and the retrospective design limit prospective interpretation. Within-region crop classification reached high accuracy across Germany, Hokkaido and Michigan, although cross-region transfer was asymmetric and usually required retraining \cite{murakami_within-_2025}.

\begin{figure}[!b]
\centering
\includegraphics[width=1.\columnwidth]{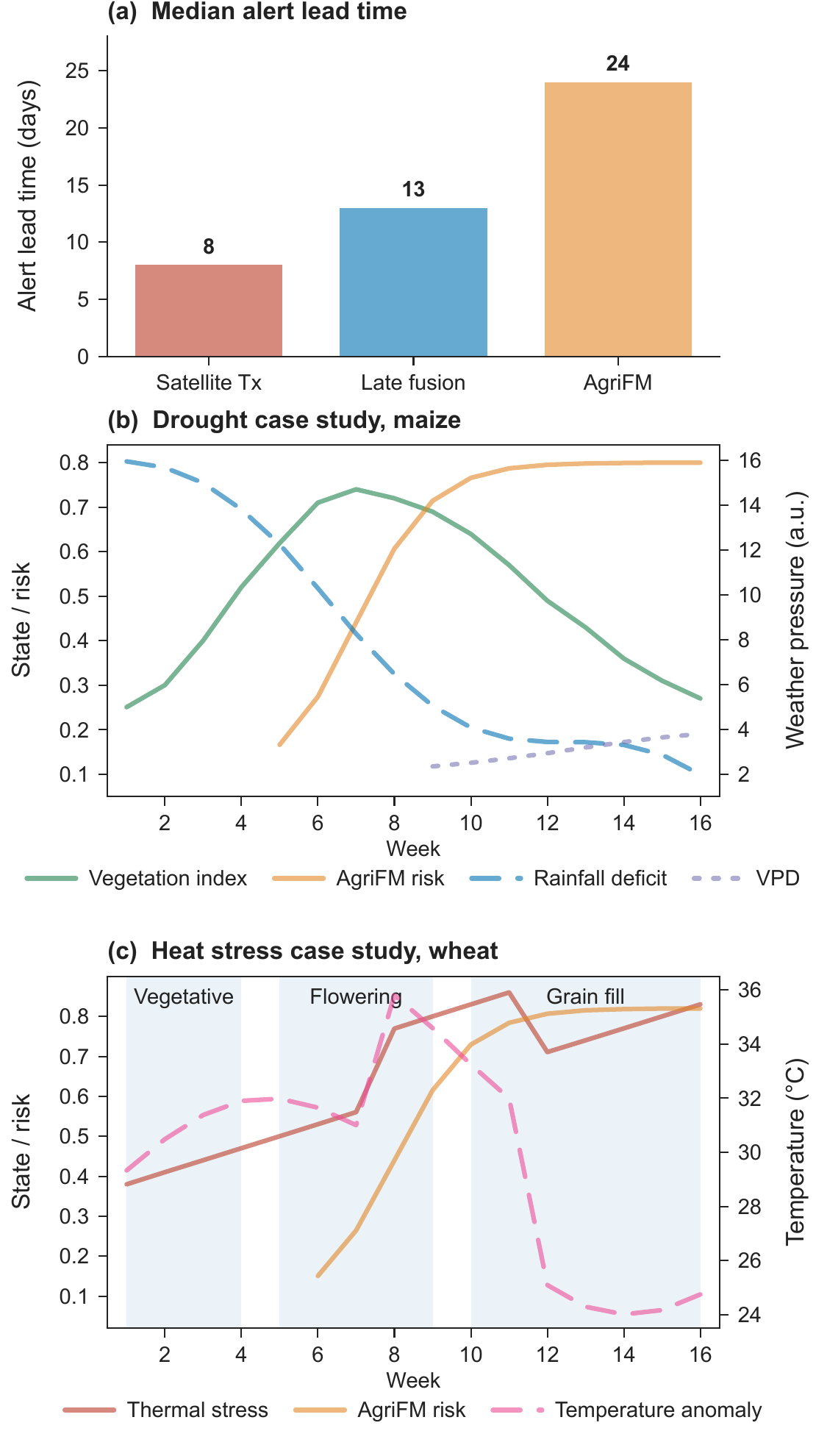}
\caption{Retrospective crop-stress early warning during extreme events. (a) Median alert lead time for a satellite-only Transformer, a late-fusion model and AgriFM. (b) Illustrative maize drought case linking vegetation condition and predicted risk with rainfall deficit and vapour-pressure deficit. (c) Illustrative wheat heat case linking thermal stress and predicted risk with temperature anomaly during the vegetative, flowering and grain-filling stages. Reprinted from \cite{albahli_multimodal_2026} under a CC BY 4.0 License.}
\label{fig:eofm_application_t3_drought}
\end{figure}

Phenology is a particularly important source of temporal structure in agricultural modelling. Spectral, thermal and radar responses change through crop development, so observations acquired at the same calendar date can represent different crop states across years, regions and management systems. Models that do not account for these shifts risk learning local crop calendars with limited transferability. Phenology-conditioned representations can align observations across planting calendars and developmental stages, improving transfer across regions and years. Phenological stage also conditions crop-stress interpretation because water or heat stress during establishment, flowering or grain filling can have different consequences for subsequent crop development and yield \cite{albahli_multimodal_2026} (Fig.~\ref{fig:eofm_application_t3_drought}).
A broader independent benchmark evaluated AlphaEarth for yield at county and field scales, tillage at county and field scales, and field-level cover-crop detection. AlphaEarth was competitive with purpose-built EO predictors under several local evaluations. Purpose-built EO predictors transferred more effectively across ecoregions and from county to field scales \cite{ma_harvesting_2026}. Under a held-out drought, latent-space drift did not reliably predict yield error and therefore could not serve as a dependable uncertainty estimate \cite{nejadshamsi_reliability_2026}.

Interpretation of these results is also constrained by the agricultural reference data available for benchmarking. Public yield records are commonly available at county or regional support, as illustrated by the county-level study of Fang et al. \cite{fang_application_2026}. Field-level yield and management records are less accessible. In the benchmark by Ma et al. \cite{ma_harvesting_2026}, field-level yield, tillage and cover-crop observations were supplied by a commercial partner, which creates an important mismatch between benchmark support and the field or sub-field scales at which many agricultural decisions are made. Consequently, strong performance against aggregated statistics does not establish equivalent skill for within-field yield, phenology, irrigation, tillage or other management decisions, and the scarcity of diverse, openly reusable field-scale labels remains an important limitation for operational benchmarking.

Retrospective AgriFM stress alerts preceded those from a satellite-only Transformer \cite{albahli_multimodal_2026}. Fig.~\ref{fig:eofm_application_t3_drought} shows median alert lead times of 8, 13 and 24 days for the satellite-only Transformer, late-fusion model and AgriFM, respectively, together with maize drought and wheat heat examples that relate predicted risk to rainfall deficit, vapour-pressure deficit and temperature anomaly. These historical cases demonstrate earlier retrospective detection under multimodal fusion. Prospective warning accuracy and intervention benefit remain untested.

\subsubsection{Ecological Traits and Composition}

Ecological applications span traits, taxa and vegetation composition. BotaCLIP improved plant, butterfly and soil-trophic-group prediction by aligning DOFA with botanical relev\'es (like plots), indicating that ecological semantics can benefit from explicit domain information \cite{cerna_botaclip_2026}. AlphaEarth and TESSERA embeddings supported label-efficient tree-species mapping; weighted F1 declined by 15\% and 9\%, respectively, under cross-year transfer, with the largest effects on rare species \cite{ball_geospatial_2026}. AlphaEarth also produced wetland vegetation maps comparable to a conventional multi-sensor workflow with less preprocessing, although smoother outputs may obscure narrow ecological boundaries \cite{ryan_streamlining_2026}. Google Satellite Embeddings fused with Landsat time series enabled national separation of planted and natural mangroves in China \cite{sun_first_2026}. This classification supports restoration stratification, while blue-carbon accumulation requires independent stock and flux evidence.

\subsubsection{Flood Inundation}

Flood inundation provides the clearest cross-model event tests. Global-event fine-tuning of TerraMind improved accuracy and precision, while U-Net retained higher recall \cite{tulbure_leveraging_2025}. Prithvi-CAFE improved held-out-site IoU by combining Prithvi adapters with a local CNN attention branch, demonstrating the contribution of complementary local detail \cite{kaushik_prithvi-complimentary_2026}. A peer-reviewed comparison across PlanetScope, Sentinel-1 and Sentinel-2 found 2--5\% variation among GFMs: Clay was strongest overall and approximately 4\% above U-Net across 19 sites, while Prithvi led on Sentinel-1 \cite{kaushik_assessing_2026}. Prithvi also adapted rapidly to two geographically distinct airborne-RGB flood events; the small event sample limits broader transfer claims \cite{polushko_flood_2026}. ZeroFlood reported stronger SAR-only segmentation with TerraMind using hydrodynamic simulations as reference labels \cite{kim_zeroflood_2026}. These studies establish event-, region- and sensor-transfer evidence for inundation extent. Flood hydraulics, antecedent storage, discharge and ecosystem recovery require additional process observations.

\subsubsection{Glacial-Lake Extent}

Glacial-lake mapping provides a spatial-transfer test in the cryosphere. TerraMind attained out-of-domain IoU 0.931 under spatially blocked evaluation \cite{abdrash_cryosentinel_2026}. Cryo-Bench further found that the strongest model varied with sensor, cryosphere task and tuning strategy \cite{kaushik_cryo-bench_2026}. These results support lake-extent delineation across spatial domains. Glacial-lake-outburst-flood early warning additionally requires information on lake evolution, dam stability, triggering processes and downstream exposure.

\subsubsection{Landslide Susceptibility}

Landslide susceptibility reinforces the value of hybrid local detail and explicit spatial transfer. Adding Clay context to a CNN increased Landslide4Sense F1 from 59.9\% to $64.5 \pm 1.8$\%, while Clay alone underperformed U-Net \cite{vu_clay-cnn_2026}. A separate AlphaEarth study remained dependent on conventional conditioning factors \cite{cheng_landslide_2026}. The available evidence therefore supports EOFM representations as contextual inputs within susceptibility workflows; event timing, runout and impact assessment require additional terrain, forcing and process information.

\subsubsection{Wildfire Burned Area and Severity}

Wildfire applications evaluate post-event burned area and spectral severity. LoRA generalised more effectively than full or decoder-only fine-tuning across 3,820 burned-area events while updating less than 1\% of parameters, with Prithvi-v2 strongest overall \cite{shibli_low-rank_2026}. TESSERA approached a strong spectral burned-area baseline in Portugal, while event-date spectral features remained preferable when suitable imagery existed \cite{silva_evaluating_2026}. Cross-continental AlphaEarth mapping showed that high overall accuracy can coexist with moderate burn-perimeter IoU \cite{seydi_deep_2025}. AlphaEarth also reconstructed cross-biome $\Delta$NBR and filled optical gaps for fire severity \cite{mahato_beyond_2026}. These studies characterise burn extent and post-event spectral expression. LFMC, ignition, fire behaviour and ecological recovery require temporally resolved observations and event-specific validation.

\subsection{Synthesis of Application-Level Evidence}
\label{sec:eofm_application_synthesis}

Taken together, the 71 application records describe several complementary routes from pretrained representations to ecohydrological outputs. Fine-tuning or PEFT was the most frequent pathway (22 records), followed by feature extraction (19) and mixed or benchmark workflows (16); fusion or similarity (8), probing or reconstruction (4), and retrieval (2) were less common. Across these pathways, EOFM representations most consistently provide spatial context, label efficiency and reduced feature engineering. Dynamic observations, meteorological forcing, LiDAR, topography and conventional catchment descriptors remain important for resolving temporal variability and subsurface or process information. Model access also shapes the evidence: analysis-ready AlphaEarth and TESSERA products favour feature-extraction and fusion studies, whereas downloadable checkpoints such as Prithvi, TerraMind, DOFA and Galileo support fine-tuning and parameter-efficient adaptation. Comparisons among models therefore remain conditional on representation access, ancillary inputs, downstream adaptation, reference data and validation design.

\begin{figure}[!t]
\centering
\includegraphics[width=1.\columnwidth]{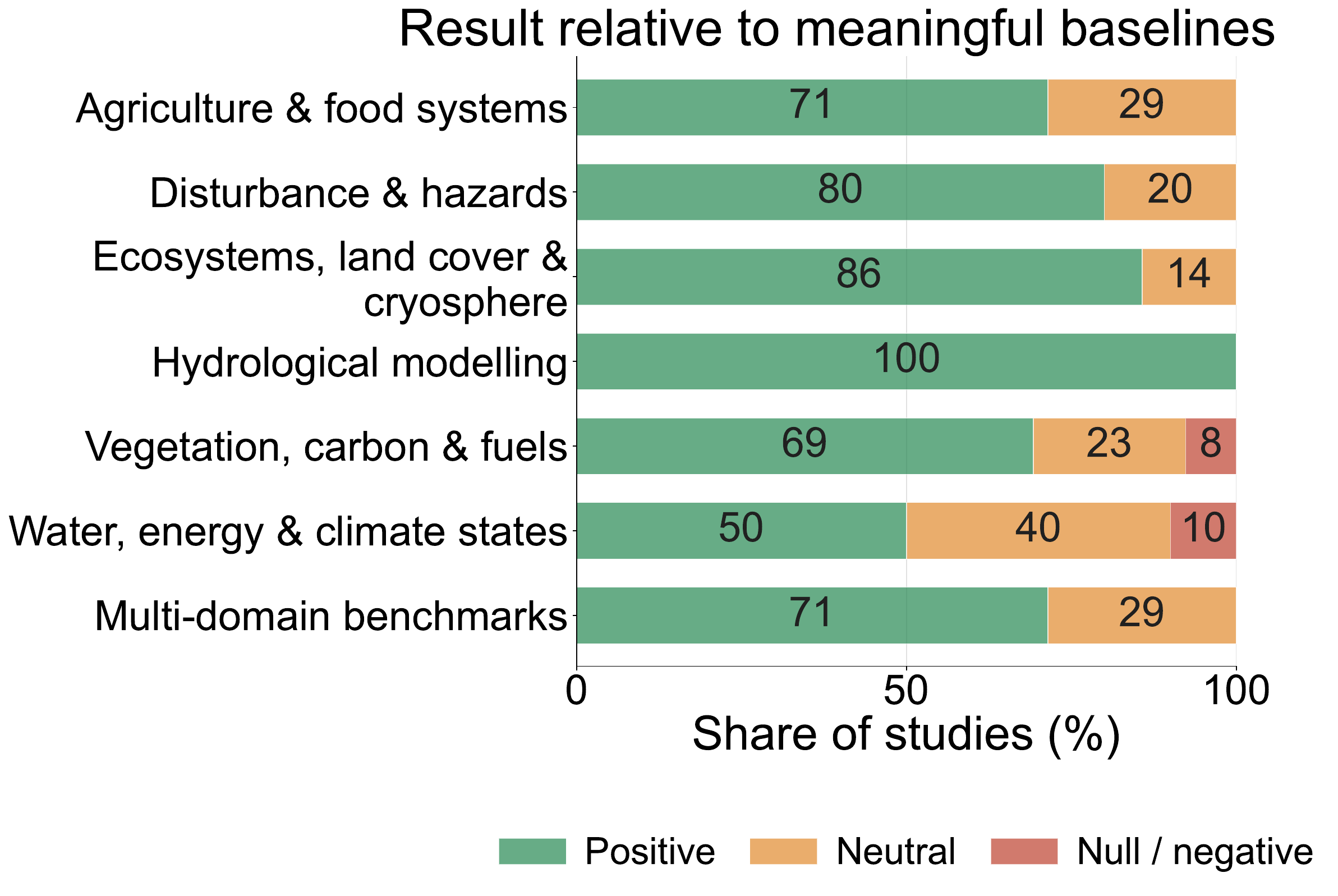}
\caption{Direction of reported results relative to meaningful baselines across the 71 eligible primary EOFM application records, grouped by application family. Stacked bars show the within-family percentage of records classified as positive, neutral (mixed or conditional), or null/negative; labels give percentages rounded to whole numbers. The categories provide a descriptive record-level synthesis across heterogeneous targets, metrics, baseline types, reference data and validation designs. They do not represent pooled effect sizes or model rankings.}
\label{fig:eofm_application_direction}
\end{figure}

Reported comparisons were favourable in most records, although their balance varied among application families (Fig.~\ref{fig:eofm_application_direction}). All five hydrological-modelling records were coded as positive, and positive results comprised 86\% of the ecosystems, land-cover and cryosphere records and 80\% of the disturbance and hazards records. Water, energy and climate states had the lowest positive share (50\%) and the largest neutral share (40\%), with 10\% classified as null or negative. Vegetation, carbon and fuels was the only other family containing a null or negative result (8\%). These proportions describe the direction of reported comparisons across unequal family sizes and heterogeneous targets, metrics, baselines, reference data and validation designs. They cannot support pooled effect estimates or general model rankings.

Evidentiary support generally decreases with inference depth. Map reproduction and sparse-label spatial regression have the broadest support, and field-referenced estimation is emerging for VWC, LFMC, SM, groundwater, water quality and SOC. Geographic transfer remains inconsistent. Flux evidence comprises a partitioned tower example for GPP, product reproduction for ET and one hybrid daily urban ET application in which meteorological forcing supplies the temporal signal. Event mapping increasingly uses region, event, sensor and temporal holdouts, whereas physical plausibility, calibrated uncertainty, causal attribution, recovery and prospective decision benefit remain sparsely evaluated. Operational readiness consequently requires evidence beyond publication status or retrospective accuracy.

Progress towards operational use requires target-matched evaluation of reference independence, temporal support, distribution-shift behaviour, physical consistency, calibrated uncertainty and implementation reproducibility at the intended spatial, temporal and decision scales. These requirements provide the basis for assessing whether an EOFM supports the claimed ecohydrological inference and whether its performance can be expected to persist beyond the original evaluation setting.

\section{Benchmarking EOFMs for Ecohydrology}
\label{sec:benchmark}

Rigorous benchmarking is essential for determining whether reported EOFM performance gains provide reliable, transferable and scientifically interpretable information for ecohydrology. This need follows directly from the preceding application synthesis, which showed that the strength of current evidence varies with inference depth, reference quality, temporal support and evaluation under environmental and sensor shifts. A credible benchmark consequently aligns the target with its spatial and temporal support, traces reference uncertainty, evaluates behaviour under relevant shifts and examines consistency with water, energy and carbon constraints \cite{karpatne_theory-guided_2017,reichstein_deep_2019,zhu_foundations_2026}. We therefore audit the diagnostic coverage of existing EOFM benchmarks and translate the recurring gaps into a target-first framework for ecohydrological evaluation.

\begin{figure}[t]
\centering
\includegraphics[width=0.98\columnwidth]{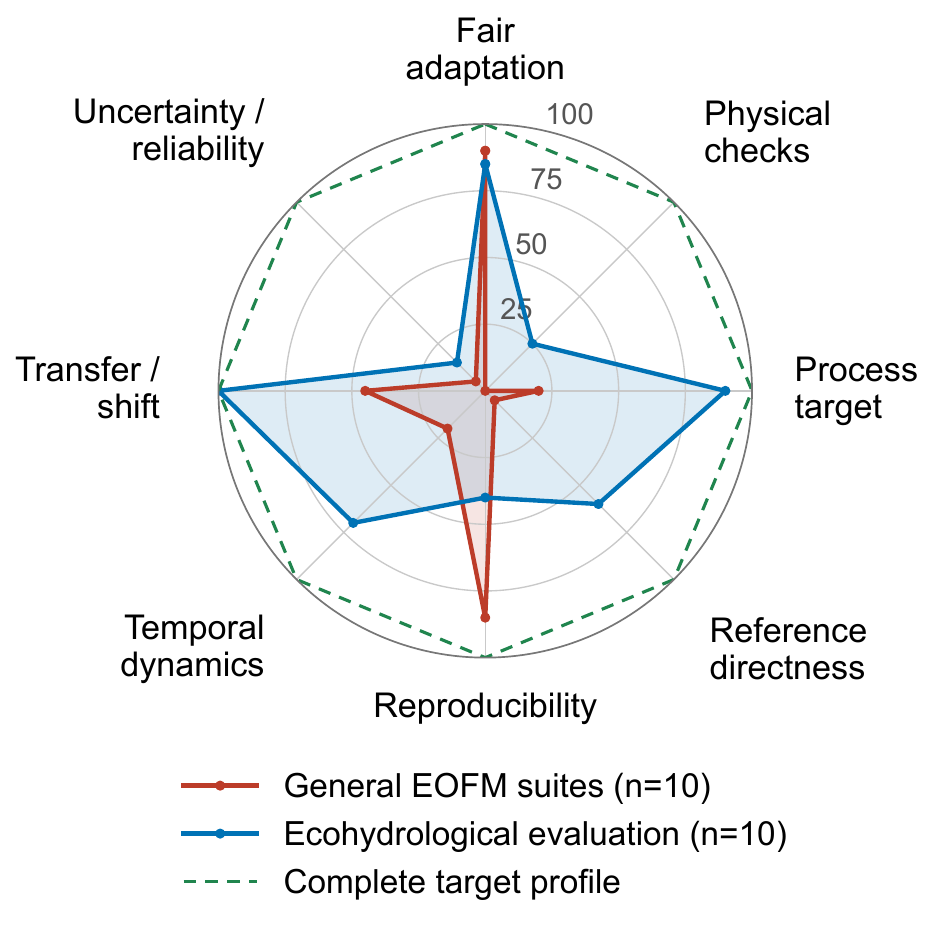}
\caption{Mean study-level benchmark coverage across eight evaluation dimensions for general EOFM suites ($n=10$) and ecohydrological evaluations ($n=10$). Ordinal scores of 0, 1 and 2 denote absent, partial or indirect, and explicit coverage, respectively (see Tables~\ref{tab:eofm_benchmark_coverage_general} and \ref{tab:eofm_benchmark_coverage_stress}), and are rescaled to 0--100\%. The dashed ring represents complete coverage across all dimensions.}
\label{fig:eofm_bench_coverage}
\end{figure}

\subsection{Evaluation Scope of Existing EOFM Benchmarks}
\label{sec:benchmark_diagnostic}

EOFM benchmarking is evolving rapidly as new model families, embedding products and evaluation suites are released. We characterise the current benchmark landscape to identify the evaluation practices already established for model comparison and the evidence available for ecohydrological inference. General suites such as GEO-Bench, PhilEO Bench, PANGAEA, GEO-Bench-2, REOBench and EarthShift provide shared protocols across tasks, regions, time periods and robustness settings \cite{lacoste_geo-bench_2023,fibaek_phileo_2024,marsocci_pangaea_2025,simumba_geo-bench-2_2026,li_reobench_2026,doerksen_earthshift_2026}. Ecohydrological evaluations extend this landscape to coupled water, energy and carbon targets.

The ecohydrological meaning of a benchmark result depends on its position within the EO inference hierarchy (Fig.~\ref{fig:01_ecohydrol_hierarchy}). T1 evaluation tests retrievals and indices against references matched to the sensing depth and spatial--temporal support; T2 evaluation addresses estimated states and fluxes through independent observations, temporal dynamics and physical relationships; T3 evaluation examines mechanisms, events and decision-relevant outcomes under environmental, temporal and sensor shifts. Therefore, we organise benchmark evidence into eight dimensions. \textit{Process target} and \textit{reference directness} establish what is inferred and which observations support the claim; \textit{temporal dynamics}, \textit{transfer or shift} and \textit{physical consistency} test the expected behaviour of ecohydrological processes; \textit{fair adaptation} identifies the contribution of the pretrained representation; \textit{uncertainty or reliability} quantifies confidence in the prediction; and \textit{reproducibility} enables independent evaluation. This framework builds on capability profiling and controlled model comparison in current EOFM research \cite{corley_no_2026,zhu_foundations_2026}.

We applied these dimensions to 20 comparative studies, with 10 each for the general EOFM benchmark suites and ecohydrological evaluations. Each dimension was coded from the reported evaluation design. A score of 0 indicates that the criterion was absent, 1 indicates coverage through a proxy, restricted subset or incomplete test, and 2 indicates that the criterion was explicitly operationalised. Group means were rescaled from 0--2 to 0--100\% for Fig.~\ref{fig:eofm_bench_coverage}; Tables~\ref{tab:eofm_benchmark_coverage_general} and \ref{tab:eofm_benchmark_coverage_stress} report the study-level scores.

\begin{table*}[t]
\centering
\begin{threeparttable}
\definecolor{NEJMBlue}{HTML}{0072B5}
\definecolor{NEJMYellow}{HTML}{FFDC91}
\colorlet{NEJMBlueAlpha}{NEJMBlue!60!white}
\colorlet{NEJMYellowAlpha}{NEJMYellow!60!white}
\caption{Record-level benchmark coverage for general EOFM suites. Benchmarks and coverage dimensions are ordered alphabetically.}
\label{tab:eofm_benchmark_coverage_general}
\renewcommand{\arraystretch}{1.12}
\setlength{\tabcolsep}{2.2pt}
\footnotesize
\newcommand{\benchmarkhead}[1]{\begin{tabular}[t]{@{}l@{}}#1\end{tabular}}
\begin{tabularx}{\textwidth}{@{}>{\raggedright\arraybackslash}p{0.24\textwidth}*{8}{>{\raggedright\arraybackslash}X}@{}}
\toprule
\benchmarkhead{Benchmark} & \benchmarkhead{Fair\\adaptation} & \benchmarkhead{Physical\\checks} & \benchmarkhead{Process\\target} & \benchmarkhead{Reference\\directness} & \benchmarkhead{Reproduc-\\ibility} & \benchmarkhead{Temporal\\dynamics} & \benchmarkhead{Transfer/\\shift} & \benchmarkhead{Uncertainty/\\reliability} \\
\midrule
Cryo-Bench \cite{kaushik_cryo-bench_2026} & \cellcolor{NEJMBlueAlpha}\makebox[\linewidth][c]{2} & \cellcolor[HTML]{FFFFFF}\makebox[\linewidth][c]{0} & \cellcolor[HTML]{FFFFFF}\makebox[\linewidth][c]{0} & \cellcolor[HTML]{FFFFFF}\makebox[\linewidth][c]{0} & \cellcolor{NEJMBlueAlpha}\makebox[\linewidth][c]{2} & \cellcolor[HTML]{FFFFFF}\makebox[\linewidth][c]{0} & \cellcolor{NEJMYellowAlpha}\makebox[\linewidth][c]{1} & \cellcolor[HTML]{FFFFFF}\makebox[\linewidth][c]{0} \\
EarthShift \cite{doerksen_earthshift_2026} & \cellcolor{NEJMBlueAlpha}\makebox[\linewidth][c]{2} & \cellcolor[HTML]{FFFFFF}\makebox[\linewidth][c]{0} & \cellcolor[HTML]{FFFFFF}\makebox[\linewidth][c]{0} & \cellcolor[HTML]{FFFFFF}\makebox[\linewidth][c]{0} & \cellcolor{NEJMBlueAlpha}\makebox[\linewidth][c]{2} & \cellcolor{NEJMYellowAlpha}\makebox[\linewidth][c]{1} & \cellcolor{NEJMBlueAlpha}\makebox[\linewidth][c]{2} & \cellcolor[HTML]{FFFFFF}\makebox[\linewidth][c]{0} \\
GEO-Bench \cite{lacoste_geo-bench_2023} & \cellcolor{NEJMBlueAlpha}\makebox[\linewidth][c]{2} & \cellcolor[HTML]{FFFFFF}\makebox[\linewidth][c]{0} & \cellcolor[HTML]{FFFFFF}\makebox[\linewidth][c]{0} & \cellcolor[HTML]{FFFFFF}\makebox[\linewidth][c]{0} & \cellcolor{NEJMBlueAlpha}\makebox[\linewidth][c]{2} & \cellcolor[HTML]{FFFFFF}\makebox[\linewidth][c]{0} & \cellcolor[HTML]{FFFFFF}\makebox[\linewidth][c]{0} & \cellcolor[HTML]{FFFFFF}\makebox[\linewidth][c]{0} \\
GEO-Bench-2 \cite{simumba_geo-bench-2_2026} & \cellcolor{NEJMBlueAlpha}\makebox[\linewidth][c]{2} & \cellcolor[HTML]{FFFFFF}\makebox[\linewidth][c]{0} & \cellcolor{NEJMYellowAlpha}\makebox[\linewidth][c]{1} & \cellcolor[HTML]{FFFFFF}\makebox[\linewidth][c]{0} & \cellcolor{NEJMBlueAlpha}\makebox[\linewidth][c]{2} & \cellcolor{NEJMYellowAlpha}\makebox[\linewidth][c]{1} & \cellcolor[HTML]{FFFFFF}\makebox[\linewidth][c]{0} & \cellcolor{NEJMYellowAlpha}\makebox[\linewidth][c]{1} \\
Landsat-Bench \cite{corley_landsat-bench_2025} & \cellcolor{NEJMBlueAlpha}\makebox[\linewidth][c]{2} & \cellcolor[HTML]{FFFFFF}\makebox[\linewidth][c]{0} & \cellcolor[HTML]{FFFFFF}\makebox[\linewidth][c]{0} & \cellcolor[HTML]{FFFFFF}\makebox[\linewidth][c]{0} & \cellcolor{NEJMBlueAlpha}\makebox[\linewidth][c]{2} & \cellcolor[HTML]{FFFFFF}\makebox[\linewidth][c]{0} & \cellcolor[HTML]{FFFFFF}\makebox[\linewidth][c]{0} & \cellcolor[HTML]{FFFFFF}\makebox[\linewidth][c]{0} \\
MMEarth-Bench \cite{gordon_mmearth-bench_2026} & \cellcolor{NEJMBlueAlpha}\makebox[\linewidth][c]{2} & \cellcolor[HTML]{FFFFFF}\makebox[\linewidth][c]{0} & \cellcolor{NEJMBlueAlpha}\makebox[\linewidth][c]{2} & \cellcolor{NEJMYellowAlpha}\makebox[\linewidth][c]{1} & \cellcolor{NEJMBlueAlpha}\makebox[\linewidth][c]{2} & \cellcolor[HTML]{FFFFFF}\makebox[\linewidth][c]{0} & \cellcolor{NEJMBlueAlpha}\makebox[\linewidth][c]{2} & \cellcolor[HTML]{FFFFFF}\makebox[\linewidth][c]{0} \\
PANGAEA \cite{marsocci_pangaea_2025} & \cellcolor{NEJMBlueAlpha}\makebox[\linewidth][c]{2} & \cellcolor[HTML]{FFFFFF}\makebox[\linewidth][c]{0} & \cellcolor{NEJMYellowAlpha}\makebox[\linewidth][c]{1} & \cellcolor[HTML]{FFFFFF}\makebox[\linewidth][c]{0} & \cellcolor{NEJMBlueAlpha}\makebox[\linewidth][c]{2} & \cellcolor{NEJMYellowAlpha}\makebox[\linewidth][c]{1} & \cellcolor{NEJMYellowAlpha}\makebox[\linewidth][c]{1} & \cellcolor[HTML]{FFFFFF}\makebox[\linewidth][c]{0} \\
PhilEO Bench \cite{fibaek_phileo_2024} & \cellcolor{NEJMBlueAlpha}\makebox[\linewidth][c]{2} & \cellcolor[HTML]{FFFFFF}\makebox[\linewidth][c]{0} & \cellcolor[HTML]{FFFFFF}\makebox[\linewidth][c]{0} & \cellcolor[HTML]{FFFFFF}\makebox[\linewidth][c]{0} & \cellcolor{NEJMBlueAlpha}\makebox[\linewidth][c]{2} & \cellcolor[HTML]{FFFFFF}\makebox[\linewidth][c]{0} & \cellcolor[HTML]{FFFFFF}\makebox[\linewidth][c]{0} & \cellcolor[HTML]{FFFFFF}\makebox[\linewidth][c]{0} \\
REOBench \cite{li_reobench_2026} & \cellcolor{NEJMYellowAlpha}\makebox[\linewidth][c]{1} & \cellcolor[HTML]{FFFFFF}\makebox[\linewidth][c]{0} & \cellcolor[HTML]{FFFFFF}\makebox[\linewidth][c]{0} & \cellcolor[HTML]{FFFFFF}\makebox[\linewidth][c]{0} & \cellcolor{NEJMYellowAlpha}\makebox[\linewidth][c]{1} & \cellcolor[HTML]{FFFFFF}\makebox[\linewidth][c]{0} & \cellcolor{NEJMBlueAlpha}\makebox[\linewidth][c]{2} & \cellcolor[HTML]{FFFFFF}\makebox[\linewidth][c]{0} \\
Satell. Imag. Robustness \cite{rotich_evaluating_2025} & \cellcolor{NEJMYellowAlpha}\makebox[\linewidth][c]{1} & \cellcolor[HTML]{FFFFFF}\makebox[\linewidth][c]{0} & \cellcolor[HTML]{FFFFFF}\makebox[\linewidth][c]{0} & \cellcolor[HTML]{FFFFFF}\makebox[\linewidth][c]{0} & \cellcolor[HTML]{FFFFFF}\makebox[\linewidth][c]{0} & \cellcolor{NEJMYellowAlpha}\makebox[\linewidth][c]{1} & \cellcolor{NEJMYellowAlpha}\makebox[\linewidth][c]{1} & \cellcolor[HTML]{FFFFFF}\makebox[\linewidth][c]{0} \\
\bottomrule
\end{tabularx}
\begin{tablenotes}[flushleft]
\scriptsize
\item[] \textit{Note:} White, NEJM yellow and NEJM blue cells denote scores 0, 1 and 2, respectively. Score 0 denotes that the criterion was not demonstrated; score 1 denotes partial or indirect coverage; and score 2 denotes explicit coverage.
\end{tablenotes}
\end{threeparttable}
\end{table*}

\begin{table*}[t]
\centering
\begin{threeparttable}
\definecolor{NEJMBlue}{HTML}{0072B5}
\definecolor{NEJMYellow}{HTML}{FFDC91}
\colorlet{NEJMBlueAlpha}{NEJMBlue!60!white}
\colorlet{NEJMYellowAlpha}{NEJMYellow!60!white}
\caption{Benchmark coverage for ecohydrological evaluations, using the dimensions and scores defined in Table~\ref{tab:eofm_benchmark_coverage_general}. Records are grouped by inference tier and ordered alphabetically within tiers.}
\label{tab:eofm_benchmark_coverage_stress}
\renewcommand{\arraystretch}{1.12}
\setlength{\tabcolsep}{1.9pt}
\footnotesize
\newcommand{\benchmarkhead}[1]{\begin{tabular}[t]{@{}l@{}}#1\end{tabular}}
\begin{tabularx}{\textwidth}{@{}>{\raggedright\arraybackslash}p{0.07\textwidth}>{\raggedright\arraybackslash}p{0.24\textwidth}*{8}{>{\raggedright\arraybackslash}X}@{}}
\toprule
\benchmarkhead{Inference\\tier} & \benchmarkhead{Ecohydrological evaluations} & \benchmarkhead{Fair\\adaptation} & \benchmarkhead{Physical\\checks} & \benchmarkhead{Process\\target} & \benchmarkhead{Reference\\directness} & \benchmarkhead{Reproduc-\\ibility} & \benchmarkhead{Temporal\\dynamics} & \benchmarkhead{Transfer/\\shift} & \benchmarkhead{Uncertainty/\\reliability} \\
\midrule
\multirow[t]{2}{0.07\textwidth}{T1} & Hydrologic Mini-JEPAs \cite{rahman_sensor-specialized_2026} & \cellcolor{NEJMBlueAlpha}\makebox[\linewidth][c]{2} & \cellcolor{NEJMYellowAlpha}\makebox[\linewidth][c]{1} & \cellcolor{NEJMBlueAlpha}\makebox[\linewidth][c]{2} & \cellcolor{NEJMYellowAlpha}\makebox[\linewidth][c]{1} & \cellcolor[HTML]{FFFFFF}\makebox[\linewidth][c]{0} & \cellcolor{NEJMYellowAlpha}\makebox[\linewidth][c]{1} & \cellcolor{NEJMBlueAlpha}\makebox[\linewidth][c]{2} & \cellcolor[HTML]{FFFFFF}\makebox[\linewidth][c]{0} \\
 & SM comparison \cite{kontogiorgakis_comparative_2026} & \cellcolor{NEJMBlueAlpha}\makebox[\linewidth][c]{2} & \cellcolor{NEJMYellowAlpha}\makebox[\linewidth][c]{1} & \cellcolor{NEJMBlueAlpha}\makebox[\linewidth][c]{2} & \cellcolor{NEJMBlueAlpha}\makebox[\linewidth][c]{2} & \cellcolor[HTML]{FFFFFF}\makebox[\linewidth][c]{0} & \cellcolor{NEJMYellowAlpha}\makebox[\linewidth][c]{1} & \cellcolor{NEJMBlueAlpha}\makebox[\linewidth][c]{2} & \cellcolor[HTML]{FFFFFF}\makebox[\linewidth][c]{0} \\
\specialrule{0.6pt}{1.5pt}{1.5pt}
\multirow[t]{3}{0.07\textwidth}{T2} & Global biomass \cite{sialelli_above-ground_2026} & \cellcolor{NEJMYellowAlpha}\makebox[\linewidth][c]{1} & \cellcolor[HTML]{FFFFFF}\makebox[\linewidth][c]{0} & \cellcolor{NEJMBlueAlpha}\makebox[\linewidth][c]{2} & \cellcolor{NEJMYellowAlpha}\makebox[\linewidth][c]{1} & \cellcolor{NEJMYellowAlpha}\makebox[\linewidth][c]{1} & \cellcolor{NEJMYellowAlpha}\makebox[\linewidth][c]{1} & \cellcolor{NEJMBlueAlpha}\makebox[\linewidth][c]{2} & \cellcolor[HTML]{FFFFFF}\makebox[\linewidth][c]{0} \\
 & Hydrological generalisation \cite{ou_foundation-scale_2026} & \cellcolor{NEJMBlueAlpha}\makebox[\linewidth][c]{2} & \cellcolor{NEJMYellowAlpha}\makebox[\linewidth][c]{1} & \cellcolor{NEJMBlueAlpha}\makebox[\linewidth][c]{2} & \cellcolor{NEJMBlueAlpha}\makebox[\linewidth][c]{2} & \cellcolor{NEJMYellowAlpha}\makebox[\linewidth][c]{1} & \cellcolor{NEJMBlueAlpha}\makebox[\linewidth][c]{2} & \cellcolor{NEJMBlueAlpha}\makebox[\linewidth][c]{2} & \cellcolor[HTML]{FFFFFF}\makebox[\linewidth][c]{0} \\
 & Ungauged basin prediction \cite{heudorfer_better_2026} & \cellcolor{NEJMBlueAlpha}\makebox[\linewidth][c]{2} & \cellcolor{NEJMYellowAlpha}\makebox[\linewidth][c]{1} & \cellcolor{NEJMBlueAlpha}\makebox[\linewidth][c]{2} & \cellcolor{NEJMBlueAlpha}\makebox[\linewidth][c]{2} & \cellcolor{NEJMBlueAlpha}\makebox[\linewidth][c]{2} & \cellcolor{NEJMBlueAlpha}\makebox[\linewidth][c]{2} & \cellcolor{NEJMBlueAlpha}\makebox[\linewidth][c]{2} & \cellcolor[HTML]{FFFFFF}\makebox[\linewidth][c]{0} \\
\specialrule{0.6pt}{1.5pt}{1.5pt}
\multirow[t]{5}{0.07\textwidth}{T3} & Agricultural reliability \cite{nejadshamsi_reliability_2026} & \cellcolor{NEJMYellowAlpha}\makebox[\linewidth][c]{1} & \cellcolor[HTML]{FFFFFF}\makebox[\linewidth][c]{0} & \cellcolor{NEJMBlueAlpha}\makebox[\linewidth][c]{2} & \cellcolor{NEJMBlueAlpha}\makebox[\linewidth][c]{2} & \cellcolor{NEJMYellowAlpha}\makebox[\linewidth][c]{1} & \cellcolor{NEJMBlueAlpha}\makebox[\linewidth][c]{2} & \cellcolor{NEJMBlueAlpha}\makebox[\linewidth][c]{2} & \cellcolor{NEJMYellowAlpha}\makebox[\linewidth][c]{1} \\
 & Agriculture transfer \cite{shang_benchmarking_2026} & \cellcolor{NEJMYellowAlpha}\makebox[\linewidth][c]{1} & \cellcolor[HTML]{FFFFFF}\makebox[\linewidth][c]{0} & \cellcolor{NEJMYellowAlpha}\makebox[\linewidth][c]{1} & \cellcolor[HTML]{FFFFFF}\makebox[\linewidth][c]{0} & \cellcolor{NEJMYellowAlpha}\makebox[\linewidth][c]{1} & \cellcolor{NEJMYellowAlpha}\makebox[\linewidth][c]{1} & \cellcolor{NEJMBlueAlpha}\makebox[\linewidth][c]{2} & \cellcolor[HTML]{FFFFFF}\makebox[\linewidth][c]{0} \\
 & AgriFM stress and yield \cite{albahli_multimodal_2026} & \cellcolor{NEJMBlueAlpha}\makebox[\linewidth][c]{2} & \cellcolor{NEJMYellowAlpha}\makebox[\linewidth][c]{1} & \cellcolor{NEJMBlueAlpha}\makebox[\linewidth][c]{2} & \cellcolor{NEJMYellowAlpha}\makebox[\linewidth][c]{1} & \cellcolor[HTML]{FFFFFF}\makebox[\linewidth][c]{0} & \cellcolor{NEJMBlueAlpha}\makebox[\linewidth][c]{2} & \cellcolor{NEJMBlueAlpha}\makebox[\linewidth][c]{2} & \cellcolor{NEJMBlueAlpha}\makebox[\linewidth][c]{2} \\
 & Flood inundation \cite{kaushik_assessing_2026} & \cellcolor{NEJMBlueAlpha}\makebox[\linewidth][c]{2} & \cellcolor[HTML]{FFFFFF}\makebox[\linewidth][c]{0} & \cellcolor{NEJMYellowAlpha}\makebox[\linewidth][c]{1} & \cellcolor[HTML]{FFFFFF}\makebox[\linewidth][c]{0} & \cellcolor{NEJMYellowAlpha}\makebox[\linewidth][c]{1} & \cellcolor{NEJMYellowAlpha}\makebox[\linewidth][c]{1} & \cellcolor{NEJMBlueAlpha}\makebox[\linewidth][c]{2} & \cellcolor[HTML]{FFFFFF}\makebox[\linewidth][c]{0} \\
 & Harvesting AlphaEarth \cite{ma_harvesting_2026} & \cellcolor{NEJMBlueAlpha}\makebox[\linewidth][c]{2} & \cellcolor[HTML]{FFFFFF}\makebox[\linewidth][c]{0} & \cellcolor{NEJMBlueAlpha}\makebox[\linewidth][c]{2} & \cellcolor{NEJMYellowAlpha}\makebox[\linewidth][c]{1} & \cellcolor{NEJMYellowAlpha}\makebox[\linewidth][c]{1} & \cellcolor{NEJMYellowAlpha}\makebox[\linewidth][c]{1} & \cellcolor{NEJMBlueAlpha}\makebox[\linewidth][c]{2} & \cellcolor[HTML]{FFFFFF}\makebox[\linewidth][c]{0} \\
\bottomrule
\end{tabularx}
\begin{tablenotes}[flushleft]
\scriptsize
\item[] \textit{Note:} T1, T2 and T3 denote retrievals and indices, estimated states and fluxes, and mechanisms, events and decision-relevant outcomes, respectively (Fig.~\ref{fig:01_ecohydrol_hierarchy}).
\end{tablenotes}
\end{threeparttable}
\end{table*}

\subsubsection{General EOFM Benchmark Suites}

General suites establish strong comparison protocols. Their mean coverage reached 90\% for fair adaptation and 85\% for reproducibility (Fig.~\ref{fig:eofm_bench_coverage}). Process targets reached 20\%, direct reference observations 5\%, uncertainty or reliability 5\% and physical consistency 0\%. MMEarth-Bench \cite{gordon_mmearth-bench_2026} provides the broadest profile in this group by combining explicit process targets, controlled adaptation, reproducible implementation and geographic transfer tests, with partial coverage of reference directness (Table~\ref{tab:eofm_benchmark_coverage_general}).

\subsubsection{Ecohydrological Evaluations}

Ecohydrological evaluations provide process-centred and shift-aware evidence. Mean coverage reached 100\% for transfer or shift, 90\% for process targets, 85\% for fair adaptation, 70\% for temporal dynamics and 60\% for reference directness (Fig.~\ref{fig:eofm_bench_coverage}). Reproducibility reached 40\%, physical consistency 25\% and uncertainty or reliability 15\%. The ungauged-basin evaluation provides the broadest profile through explicit process targets, direct references, temporal dynamics, spatial transfer, fair adaptation and reproducibility, together with partial physical checks \cite{heudorfer_better_2026}. AgriFM adds explicit reliability evaluation to temporal and transfer tests for crop stress and yield \cite{albahli_multimodal_2026}. Table~\ref{tab:eofm_benchmark_coverage_stress} locates these evaluations across the inference tiers.

Together, the two groups show that benchmark strengths are distributed unevenly across the eight dimensions. Their complementary coverage and shared diagnostic gaps motivate the priorities for future benchmark development.

\subsection{Priorities for Future Ecohydrological Benchmarks}

The coverage analysis supports a three-part agenda for future EOFM benchmarks in ecohydrology. Table~\ref{tab:eofm_ecohydrology_benchmark_requirements} maps the eight dimensions in Fig.~\ref{fig:eofm_bench_coverage} to current patterns and evidence-linked priorities. We organise these priorities into established practices for standardisation, complementary strengths for integration and diagnostic gaps for priority development.

\begin{table*}[t]
\centering
\caption{Priorities for future EOFM benchmarks in ecohydrology.}
\label{tab:eofm_ecohydrology_benchmark_requirements}
\renewcommand{\arraystretch}{1.05}
\setlength{\tabcolsep}{2.5pt}
\footnotesize
\begin{tabularx}{\textwidth}{@{}>{\raggedright\arraybackslash}p{0.10\textwidth}>{\centering\arraybackslash}p{0.15\textwidth}>{\raggedright\arraybackslash}p{0.30\textwidth}>{\raggedright\arraybackslash}X@{}}
\toprule
Dimension & Coverage (\%)\newline General / Ecohydrol. & Current benchmark pattern and evidence & Future priority and demonstrated examples \\
\midrule
Fair adaptation & 90 / 85 & Shared adaptation protocols and controlled comparisons are established in general suites and selected ecohydrological evaluations \cite{lacoste_geo-bench_2023,marsocci_pangaea_2025,heudorfer_better_2026}. & Match encoder access, inputs, downstream heads, label and optimisation budgets, and repeated-run reporting. Examples include shared-protocol and factorial comparisons \cite{lacoste_geo-bench_2023,heudorfer_better_2026}. \\
Physical checks & 0 / 25 & Partial checks address SM plausibility, hydrological regimes and sensor--variable coherence \cite{kontogiorgakis_comparative_2026,ou_foundation-scale_2026,rahman_sensor-specialized_2026}. & Evaluate physical bounds, water--energy--carbon closure, partition consistency, cross-variable coherence, perturbation responses and regime-conditioned residuals. Examples demonstrate selected components of these tests \cite{kontogiorgakis_comparative_2026,ou_foundation-scale_2026,rahman_sensor-specialized_2026}. \\
Process target & 20 / 90 & Continuous soil, biomass, streamflow, crop-stress and yield targets extend evaluation across T1--T3 \cite{gordon_mmearth-bench_2026,sialelli_above-ground_2026,albahli_multimodal_2026}. & Define the physical quantity, units, inference tier and spatial--temporal support; distinguish signals, retrievals, states, fluxes, stocks, events and decisions. Examples span continuous environmental and event targets \cite{gordon_mmearth-bench_2026,sialelli_above-ground_2026,albahli_multimodal_2026}. \\
Reference directness & 5 / 60 & General suites primarily use product and mapped references \cite{lacoste_geo-bench_2023,gordon_mmearth-bench_2026,marsocci_pangaea_2025}. Ecohydrological studies also use station and gauge observations \cite{kontogiorgakis_comparative_2026,ou_foundation-scale_2026,heudorfer_better_2026}. & Pair product agreement with matched \emph{in situ}, field, flux-tower, gauge or inventory observations, and report observation and representation errors separately. Examples use station and gauge references \cite{kontogiorgakis_comparative_2026,heudorfer_better_2026}. \\
Reproducibility & 85 / 40 & Open benchmark assets, fixed splits and executable protocols are common in general suites and present in selected ecohydrological evaluations \cite{lacoste_geo-bench_2023,marsocci_pangaea_2025,heudorfer_better_2026}. & Release versioned data, splits, preprocessing, code, weights or embeddings, hyperparameters, compute requirements and licences. Examples provide reusable benchmark or evaluation resources \cite{lacoste_geo-bench_2023,marsocci_pangaea_2025,heudorfer_better_2026}. \\
Temporal dynamics & 20 / 70 & Current tests examine paired temporal shifts, temporal generalisation and event development \cite{doerksen_earthshift_2026,ou_foundation-scale_2026,albahli_multimodal_2026}. & Test seasonal phase, event onset, thresholds, antecedent memory, lag and recovery at process-appropriate cadence. Examples evaluate temporal generalisation and event evolution \cite{ou_foundation-scale_2026,albahli_multimodal_2026}. \\
Transfer/shift & 45 / 100 & Explicit tests address geographic, temporal, sensor, scale and robustness shifts \cite{doerksen_earthshift_2026,li_reobench_2026,shang_benchmarking_2026}. & Use explicit site-, basin-, biome-, year-, sensor-, scale-, regime- and extreme-event holdouts, and report performance changes from the in-domain setting. Examples span real-world and corruption shifts \cite{doerksen_earthshift_2026,li_reobench_2026,shang_benchmarking_2026}. \\
Uncertainty/\newline reliability & 5 / 15 & GEO-Bench-2 includes an uncertainty-related task component, and crop-stress and yield studies contribute drift and calibrated reliability analyses \cite{simumba_geo-bench-2_2026,nejadshamsi_reliability_2026,albahli_multimodal_2026}. & Report calibration, interval coverage, sharpness, out-of-domain warning and decomposition of observation, parameter and structural uncertainty under in-domain and shifted conditions. Examples test drift and reliability \cite{nejadshamsi_reliability_2026,albahli_multimodal_2026}. \\
\bottomrule
\end{tabularx}
\end{table*}

\subsubsection{Established Practices for Standardisation}

Fair adaptation is the most consistently established dimension, reaching 90\% coverage in general suites and 85\% in ecohydrological evaluations. General suites also reach 85\% coverage for reproducibility through shared protocols, fixed splits and open benchmark assets. Future benchmarks can consolidate these practices through common protocols that align encoder access, inputs, downstream heads, label budgets, optimisation budgets and repeated-run reporting. For served embeddings, comparable evaluation also requires matched spatial and temporal support, pooling, update frequency, access and licence \cite{stewart_earth_2026}.

\subsubsection{Complementary Strengths for Integration}

Coverage in general suites and ecohydrological evaluations, respectively, differs most for process targets (20\% and 90\%), reference directness (5\% and 60\%), temporal dynamics (20\% and 70\%), transfer or shift (45\% and 100\%) and reproducibility (85\% and 40\%). These complementary profiles define the main integration task for future benchmarks: combine controlled comparison and reproducible implementation with process-centred targets, direct reference evidence and shift-aware evaluation. A target-matched benchmark begins by defining the physical quantity, units, inference tier and spatial and temporal support. Product-based comparisons and direct observations provide complementary reference layers and should be reported separately. Temporal tests should resolve seasonal phase, event onset, antecedent memory, lag and recovery. Shift tests should hold out relevant sites, basins, biomes, years, sensors, scales, regimes and extremes.

\subsubsection{Diagnostic Gaps for Priority Development}

Physical checks remain limited, with coverage of 0\% in general suites and 25\% in ecohydrological evaluations. Uncertainty or reliability is similarly underdeveloped, at 5\% and 15\%, respectively. Future evaluation should connect predictive accuracy with physical bounds and water--energy--carbon closure, calibrated uncertainty and reproducible implementation. Raw EO, engineered predictors, randomly initialised models, task-specific models, process-based models and hybrid workflows provide the baselines needed to attribute gains to pretraining. Null and negative results against these baselines identify the conditions under which reusable representations add transferable ecohydrological information \cite{kontogiorgakis_comparative_2026,ma_harvesting_2026}.

Together, these priorities convert the coverage assessment into a practical design for future benchmark development. Their implementation depends on harmonised and diverse data infrastructure that preserves observation provenance, spatial and temporal support, versioned evaluation splits and access to the evaluated models or embeddings.

\subsection{Multiscale Data Infrastructure for Ecohydrological Benchmarking}

Ecohydrological benchmarking can draw on two complementary forms of data infrastructure. Established resources for process and model evaluation provide long-term observations, forcing data and derived products for coupled water--energy--carbon dynamics. Emerging resources extend this foundation towards EOFM evaluation through harmonised multimodal data and EO products linked to ecosystem observations. Their integration connects EOFM assessment with process evidence across spatial and temporal scales.

\subsubsection{Established Data Infrastructure for Model Evaluation}

The global eddy covariance network provides half-hourly to decadal measurements of ecosystem carbon dioxide, water-vapour and energy exchange, and PLUMBER2 harmonises 1,040 site-years from 170 flux towers for land-model evaluation \cite{xiao_insights_2026,ukkola_flux_2022}. The STEMMUS-SCOPE dataset extends site-scale evaluation with model-derived hydrological, photosynthetic and radiative variables evaluated against PLUMBER2 fluxes and FLUXNET2015 soil moisture. FluxHourly integrates simulations, eddy covariance observations, EO, meteorological data and machine learning to provide global hourly 9-km water--energy--carbon estimates for 2000--2020 \cite{wang_physically_2025,han_fluxhourly_2025}. GeoNEX Coincident Ground Observations collocates 5--10-min geostationary EO with 1,586 network sites across the Americas, enabling evaluation of diurnal surface-energy dynamics and observed carbon and water fluxes \cite{hashimoto_subsets_2026}. Together, these resources provide observational, model-derived and upscaled references whose provenance, spatial--temporal support and uncertainty define their use in EOFM evaluation.

A major constraint on agricultural benchmarking is the scarcity of public, benchmark-ready reference data at the field and sub-field scales relevant to many management decisions. Recent yield benchmarks continue to rely heavily on aggregated statistics. Fang et al. \cite{fang_application_2026} evaluated AlphaEarth using county-level USDA yield records; CropClimateX provides county-level yield and farm-management variables, including planted area, harvested area and production \cite{hohl_large-scale_2026}. Field-scale yield and management observations exist, although access remains limited: Ma et al. \cite{ma_harvesting_2026} used field-level yield, tillage and cover-crop observations supplied by Corteva Agriscience, noted that comparable records are difficult to acquire because of privacy concerns. The resulting shortage of diverse, openly reusable field-scale labels limits benchmarking of phenology, tillage, irrigation and yield at the spatial and temporal support relevant to on-farm decisions.

\subsubsection{Emerging Data Infrastructure for EOFM Evaluation}

Recent datasets organise Earth-system information specifically for FM workflows. WorldTensor aligns hundreds of environmental and socioeconomic variables on a common $0.25^{\circ}$ annual grid for planetary-scale training and evaluation \cite{rodriguez-pardo_harmonised_2026}. FluxCubes30 provides 30-m GPP and EVI cubes, pixel-level uncertainty estimates and 7-km by 7-km cutouts for 404 flux-tower-centred landscapes from 1999 to 2025 \cite{luo_fluxcubes30_2026}. Cross-calibrated Landsat observations and tower-footprint matching support evaluation of long-term carbon dynamics and fine-scale disturbance or irrigation signals. These datasets serve distinct roles in EOFM evaluation. WorldTensor provides harmonised planetary context for multimodal representation learning, and FluxCubes30 provides process-focused, tower-centred EO data for evaluating landscape-scale carbon dynamics.

In summary, ecohydrological data infrastructure for EOFM benchmarking can combine established model-evaluation resources with emerging EOFM-oriented datasets. Harmonised variables, matched spatial--temporal support, traceable target provenance and uncertainty information enable reproducible integration and preserve the evidentiary role of each dataset.

\section{Discussion and Prospects}
\label{sec:discussion}
\label{sec:prospects}

Together, Sections~\ref{sec:ecohydrol_inference}, \ref{sec:eofm_meta}, \ref{sec:eofm_capabilities} and \ref{sec:benchmark} address RQ1--RQ4, respectively, through a coherent analytical sequence that defines the ecohydrological inference problem, maps the EOFM design landscape, synthesises application-level evidence and translates recurring evidence gaps into requirements for target-first benchmarking. We now revisit these four RQs to critically examine the scientific contributions, evidentiary boundaries and interpretive limitations revealed by each analysis and to develop literature-supported prospects for future research, evaluation and scientific use.

\subsection{Observability, Inference Contracts and Uncertainty}

Revisiting RQ1, the hierarchy in Section~\ref{sec:ecohydrol_inference} shows that EO signals and complementary data support inference from calibrated measurements to ecohydrological properties, states, fluxes, mechanisms and decision-relevant outcomes through target-specific pathways. The scientific validity of these pathways depends on observation physics, reference uncertainty, scale and process constraints, which define the observability limits of every EOFM claim. Reflected-solar, thermal, microwave and fluorescence measurements provide complementary sensitivities to vegetation structure, LST, water status and photosynthetic activity; their conversion into latent states or fluxes introduces atmospheric-correction, radiative-transfer, inversion, forcing and partitioning assumptions, and predictive agreement therefore provides incomplete evidence of process fidelity \cite{jiao_remote_2026,mohammed_remote_2019,reichstein_deep_2019,shen_differentiable_2023}. Nominal pixel size describes sampling, while process support also depends on sensing depth, footprint and aggregation; comparisons among point SM probes, satellite footprints, eddy-covariance footprints and catchment responses consequently require explicit scale matching and reference-error characterisation \cite{crow_upscaling_2012,gruber_validation_2020,dorigo_international_2021}. Temporal support imposes a parallel constraint: high-frequency geostationary observations can resolve diurnal and rapidly evolving ecosystem dynamics \cite{khan_reviews_2021,yu_solar_2024}, whereas the denser sampling of TROPOMI SIF remains conditioned by signal-to-noise and spatial aggregation over small or fragmented ecosystems \cite{guanter_troposif_2021}. Canopy near-infrared reflectance provides an optical constraint on GPP \cite{badgley_canopy_2017}; physiological regulation remains an additional inference. Because cross-level feedbacks and processing choices alter the evidentiary role of a variable, tier assignment must remain target- and use-dependent, with EOFM performance interpreted through the complete observation--representation--reference chain.

Future EOFM studies can operationalise these limits by pairing a target-specific inference contract with a quantitative uncertainty budget. The contract would record the physical quantity and units, observation pathway, processing level, spatial and temporal support, reference directness and independence, scale-matching operator, governing assumptions and intended interpretation; the accompanying budget would preserve identifiable measurement, retrieval, forcing, label, sampling, parameter, structural and distributional uncertainties as information moves from observations to model outputs. This design is consistent with data-assimilation research that identifies observation-operator specification, model--data inconsistency, parameter identifiability and spatiotemporal heterogeneity as central controls on uncertainty \cite{raoult_parameter_2025}. Each scientific claim should also carry a prespecified falsification test: retrieval claims can be challenged across sensors and processing versions and against independent, error-characterised references; state and flux claims through site-, biome-, season- and scale-held-out evaluation; and mechanistic claims through perturbation responses, temporal ordering, closure constraints and, where available, intervention or counterfactual evidence \cite{karpatne_theory-guided_2017,reichstein_deep_2019,shen_differentiable_2023}. Recent tests of learned observation operators further show that acceptable prediction error can coexist with physically implausible Jacobians, motivating sensitivity diagnostics alongside output metrics \cite{shan_importance_2026}. Ensembles, hierarchical error models, perturbation experiments and cross-scale validation could then propagate the uncertainty budget and identify the conditions under which an EOFM inference remains scientifically defensible.

\subsection{Representation Capacity and Trustworthiness}
Revisiting RQ2, the evidence map in Section~\ref{sec:eofm_meta} shows that current EOFMs cover a broad yet uneven combination of EO modalities, spatial and temporal supports, representation strategies and adaptation routes. Reflected optical observations dominate, SAR provides most microwave coverage, explicit thermal pre-training remains uncommon, and passive-microwave emission and SIF are absent as explicit pre-training pathways in the eligible release set. This distribution leaves several observation pathways that are important for ecohydrological water, energy and carbon inference underrepresented, consistent with wider assessments of the sensor, scale and temporal capabilities required of Earth FMs \cite{xiao_foundation_2025,hong_foundation_2026,zhu_foundations_2026}. Representation form further conditions scientific use: runnable checkpoints enable linear probing, parameter-efficient adaptation or full fine-tuning but require accessible weights, executable code and sufficient compute; served Earth embeddings reduce downstream processing costs but retain only the information preserved by their pre-training objectives, spatial or temporal pooling and update schedules \cite{klemmer_earth_2025,stewart_earth_2026}; and harmonised tensors provide broad environmental context while their common annual grid suppresses native support differences, local gradients and event dynamics \cite{rodriguez-pardo_harmonised_2026}. The map itself is descriptive: related releases share training data, architectures, code and benchmarks, public documentation is uneven, and release counts, parameter counts and nominal resolution cannot establish process information, effective support or transfer. RQ2 is therefore answered conditionally: current EOFMs offer diverse access and adaptation routes, while event-resolving temporal support, physically complementary non-optical pathways, transparent representation lineage and empirical evidence of retained process information remain limited.

Trustworthiness should consequently be treated as a primary scientific design requirement for EOFMs and distributed Earth representations. Current perspectives identify robustness, calibrated uncertainty, interpretability, reproducibility, efficient adaptation and human- and environment-centred use as foundational properties of scientifically useful Earth FMs \cite{zhu_foundations_2026,gawlikowski_survey_2023}. The operational gap is already measurable: among 89 geospatial FMs, 50.6\% lacked accessible weights or provided only minimally guided weights, 6.7\% offered an API or graphical interface, and the 63 accessible models overwhelmingly documented benchmarks while documented uncertainty quantification was absent and explainability and interactive validation were rare \cite{young_how_2026}. Robustness evidence also shows strong task and model dependence: across 6 EO tasks and 12 image corruptions, REOBench recorded performance degradation ranging from less than 1\% to more than 20\% \cite{li_reobench_2026}. Building on emerging Earth-embedding interfaces and provenance proposals \cite{klemmer_earth_2025,stewart_earth_2026}, future work should establish a persistent, versioned registry and ``Embedding Cards'' that record source products and processing levels, observation pathways, spatial and temporal support, pooling and update cadence, masks, licences, compute requirements, data overlap, uncertainty and documented failure conditions. Each vector or prediction should remain linked to its source observations, acquisition dates, quality masks, model version and downstream adapter, with calibrated uncertainty and out-of-distribution indicators available for human review. Readiness should require independent reconstruction; identification of evidentiary support and unsupported conditions; matched-budget comparisons with raw and engineered predictors; and stable, physically credible performance under geographic, temporal, sensor, corruption and missing-modality shifts. These tests would make trustworthiness auditable and falsifiable across the full representation lifecycle.

\subsection{Ecohydrological Applications and Process-Targeted Representations}
Revisiting RQ3, the application synthesis in Section~\ref{sec:eofm_capabilities} shows that current EOFMs have most consistently demonstrated reusable contextual representation, label-efficient adaptation and multimodal fusion across retrievals, estimated states and fluxes, and event- or decision-related targets. Their contribution depends on how embeddings interact with dynamic EO, meteorological forcing, local observations and process information. This division is clearest in hybrid studies: combining AlphaEarth spatial context with 5-min GOES-18 observations reduced leave-one-station-out LST RMSE across 53 Hawai`i stations from 3.11 to 2.87~K, with the thermal observations retaining the sub-hourly trajectory \cite{lee-burkhart_geospatial_2026}; and a Prithvi--MERRA-2 model evaluated at 37 flux towers increased mean leave-one-year-out GPP $R^2$ from 0.75 for a same-input ResNet to 0.81, although meteorological forcing, NEE partitioning and tower-footprint mismatch constrain attribution of the gain \cite{szwarcman_prithvi-eo-20_2026}. Null and negative results delimit this capability: Prithvi embeddings produced negligible improvement over engineered optical, SAR and ERA5 predictors for SM at 113 International Soil Moisture Network (ISMN) stations ($R^2=0.515$ versus 0.514) \cite{kontogiorgakis_comparative_2026}; spectral indices consistently exceeded AlphaEarth embeddings for field-referenced biomass in one regenerating Andean forest \cite{rojas-lucero_spectral_2026}; and AlphaEarth was competitive for locally trained agricultural tasks but showed weaker spatial transfer, temporal sensitivity and interpretability than purpose-built EO models \cite{ma_harvesting_2026}. Evidence is strongest when the target is observable, the reference is sufficiently independent, spatial and temporal supports are matched, and environmental or sensor shifts are held out. RQ3 is therefore answered at the level of conditional capability: EOFMs can supply spatial context, reduce feature construction and support selected low-label or transfer settings; broad recovery of ecohydrological mechanisms, extreme-event dynamics and prospective decision benefit remains unestablished.

Future progress should centre on process-targeted representations whose inputs, outputs and validation expose the source of each inferred capability. The Bidirectional Encoder Representations from Transformers for Hydrology (BERTH) framework provides an emerging example by translating optical reflectance, ERA5-Land forcing and terrain into daily 30-m estimates of ET, precipitation, SM and runoff through a shared temporal architecture \cite{yao_transforming_2026}. Its coupled output space creates a basis for water-cycle analysis, while product-based pre-training, forcing dependence, unequal native supports, unavailable in situ runoff supervision and an annual runoff-closure correction delimit claims of fine-scale process recovery. TabPFN offers a distinct low-data adaptation route for structured observations: its prior-data-fitted design supports regression and uncertainty estimation for small tabular datasets \cite{hollmann_accurate_2025}, and an AlphaEarth--TabPFN workflow predicted groundwater depth from 87 wells in the Leizhou Peninsula with $R^2=0.793$ \cite{li_new_2026}. Evaluation within one hydrogeological setting leaves regional transfer unresolved, and the physical meaning of the prediction continues to depend on the embedding inputs and field references. A literature-supported research agenda should integrate dynamic optical, thermal, active and passive microwave and SIF-sensitive observations with meteorological forcing, terrain and direct networks of stream gauges, groundwater wells, soil profiles, flux towers, inventories and field campaigns, while retaining distinct variables and units for surface and root-zone SM, ET and transpiration, precipitation, runoff, groundwater, GPP and respiration. Constrained translators, differentiable process modules and data assimilation can encode water, energy and carbon relations and propagate observation and model uncertainty \cite{reichstein_deep_2019,shen_differentiable_2023,raoult_parameter_2025}. Prospective multisite and multiyear studies should compare raw EO, engineered predictors, task-specific models, frozen embeddings, fine-tuned encoders, process models and hybrid systems under matched label and adaptation budgets; use direct, scale-matched references; and test held-out regions, basins, years, extremes, sensors and missing modalities. Readiness requires calibrated uncertainty, independently supported spatial detail, credible event timing and recovery, and coherent water balance, energy closure, storage--discharge and GPP--ET behaviour under the intended environmental, temporal, sensor and scale shifts.

\subsection{Target-First Benchmarking and Future Development}

Revisiting RQ4, the audit in Section~\ref{sec:benchmark} shows that benchmarking is an act of scientific specification: it connects a model output to an ecohydrological claim through an observation and reference pathway and within a defined domain of validity. Each evaluation should therefore preserve the physical identity of the target, including its quantity, units and inference tier; its observational identity, including sensor, processing level, reference directness and spatial--temporal support; and its experimental identity, including the adaptation regime, baselines, shift design, uncertainty treatment and reproducibility conditions. PANGAEA and GEO-Bench-2 show that measured capability varies with task, modality, resolution, temporal structure and adaptation strategy; REOBench and EarthShift further distinguish sensitivity to input corruption from generalisation across geography, time, spatial resolution, sensor and dataset composition \cite{marsocci_pangaea_2026,simumba_geo-bench-2_2026,li_reobench_2026,doerksen_earthshift_2026}. These perturbations probe different failure mechanisms in ecohydrological inference and require separate diagnostic experiments. RQ4 is consequently answered through a claim-conditioned evidence profile encompassing predictive accuracy, temporal behaviour, physical consistency, uncertainty calibration, transfer across the intended domain, computational and adaptation cost, and reproducibility. Physical consistency refers to agreement with admissible variable bounds, water--energy--carbon balances, coupled-variable relations and process-relevant event timing, evaluated as falsifiable out-of-sample diagnostics \cite{karpatne_theory-guided_2017,reichstein_deep_2019,zhu_foundations_2026}. The audit also establishes a provenance requirement for interpretation: heterogeneity among records, incomplete pre-training disclosure, repeated labels, dependence among model families and uneven test severity define the uncertainty in cross-study attribution, and each benchmark result should remain linked to its data lineage, model version and complete evaluation protocol \cite{corley_no_2026}.

Together, these requirements define a staged agenda for benchmark development. Near-term priorities establish comparable evidence, medium-term priorities extend process and shift diagnostics, and long-term priorities sustain versioned evaluation as data, models and use contexts evolve.

\subsubsection{Near Term: Comparable Protocols}

Near-term work should make each comparison reproducible. Corley et al. call for shared core evaluations, reusable weights, variance reporting, common evaluation harnesses and controls for data, architecture and algorithm effects \cite{corley_no_2026}. PANGAEA and REOBench add stratified subsets, controlled runs, versioned metadata and maintained releases \cite{marsocci_pangaea_2026,li_reobench_2026}. Each ecohydrological record should state the target and units, inference tier, observation and reference pathways, spatial and temporal support, split design and pre-training overlap, model version, adaptation and compute budgets, baselines, uncertainty, and access conditions \cite{stewart_earth_2026}. This minimum record supports attribution of performance to the representation and its adaptation.

\subsubsection{Medium Term: Process and Shift Coverage}

Medium-term work should expand process and shift coverage. PANGAEA, REOBench, GEO-Bench-2 and EarthShift prioritise additional sensors, tasks, regions, uncertainty evaluation and real-world shifts \cite{marsocci_pangaea_2026,li_reobench_2026,simumba_geo-bench-2_2026,doerksen_earthshift_2026}. The ecohydrological studies audited here already cover soil moisture, biomass, streamflow, crop stress and flood inundation \cite{kontogiorgakis_comparative_2026,sialelli_above-ground_2026,heudorfer_better_2026,albahli_multimodal_2026,kaushik_assessing_2026}. Flux networks and land-model evaluation datasets provide references for coupled water, energy and carbon exchange \cite{xiao_insights_2026,ukkola_flux_2022}. These evidence streams support target-specific diagnostics of soil-water depth and support, surface-flux partitioning and closure, catchment timing and water balance, and vegetation-stress or disturbance onset and recovery. Calibrated uncertainty and separate geographic, temporal, sensor, scale, regime, extreme-event and missing-modality tests can locate performance changes within defined processes and evaluation conditions.

\subsubsection{Long Term: Persistent Evaluation}

Long-term work should convert periodic benchmark releases into a maintained evaluation programme that tracks evidence as models, data and environmental conditions evolve. EarthShift frames benchmarking as a living testbed; REOBench and PANGAEA provide versioning, preservation and extensible protocols; and Corley et al. call for rerunnable evaluations indexed by task, sensor, region and date \cite{doerksen_earthshift_2026,li_reobench_2026,marsocci_pangaea_2026,corley_no_2026}. Persistent release identifiers and frozen historical tests can preserve comparability across versions, with prospective additions of sites, years, sensors and extreme events expanding the tested domain. Linking each result to its target definition, reference provenance, uncertainty and correction history would preserve the conditions under which the evidence was obtained and support cumulative assessment for scientific and operational use. Routine operations within this programme, including data discovery and standardised test execution, may be assisted by FMs with traceable provenance and expert-reviewed acceptance criteria \cite{zhang_foundation_2025}.

\section{Conclusion}
\label{sec:conclusion}

The scientific value of EOFMs for terrestrial ecohydrology depends on whether reusable representations learned from EO data support credible process inference. In this review, we developed an observation-to-inference framework and applied it to a meta-analysis of EOFM designs, a synthesis of ecohydrological applications and an audit of benchmark studies. By linking sensing pathways, spatial and temporal support, process timescales and reference uncertainty to each target, we identified the demonstrated capabilities of current EOFMs, the conditions governing their scientific interpretation and the requirements for inference of coupled water, energy and carbon dynamics.

We progressively define and demonstrate the current evidence for translating EOFM representations into ecohydrological process inference. Firstly, the observation-to-inference hierarchy shows that EOFM relevance depends on target-specific sensing pathways, spatial and temporal support, and traceable uncertainty. Secondly, the meta-analysis reveals that pretraining remains dominated by reflected optical and active-microwave data, with sparse thermal coverage and no explicit passive-microwave-emission or SIF sensing data sources. Thirdly, the application synthesis identifies the strongest support for spatial context, label-efficient adaptation and hybrid workflows; evidence weakens with inference depth, and independent validation of fluxes, coupled dynamics, event trajectories, calibrated uncertainty and decision benefit remains sparse. Fourthly, the benchmark audit shows that general EOFM suites more often address fair adaptation and reproducibility, whereas ecohydrological evaluations more often address process targets, direct reference evidence and distribution shifts; physical consistency and uncertainty remain weakly covered across both groups.

Future progress requires a process-aware agenda aligned with inferred variables, observation pathways and process timescales. Core elements include target-specific inference contracts and uncertainty budgets, supported by traceable provenance, calibrated uncertainty and documented failure conditions. Process-targeted hybrids integrating EO, meteorological forcing, in situ data and process constraints can preserve event dynamics and water--energy--carbon relations. Persistent benchmarks linking direct references and physical diagnostics with matched adaptation and distribution-shift tests would provide the corresponding evaluation structure. Together, these developments would establish EOFMs as testable components of ecohydrological monitoring and process inference.

\section{Code and Data Availability}

The collated resources will be distributed through \url{https://github.com/yuyi13/EOFM4EcoHydrol}.

\IEEEtriggeratref{184}
\IEEEtriggercmd{\balance}
\bibliographystyle{IEEEtran}
\bibliography{references_cleaned}

\end{document}